\documentclass[10pt,journal,final,twocolumn]{IEEEtran}

\usepackage{graphicx,times,amsmath,amssymb,cite,cases,mathrsfs,booktabs,multirow,psfrag,subfigure,url,stfloats,threeparttable}
\usepackage{algorithm,algorithmic,amsthm,ulem,latexsym,bm}
\usepackage{dsfont}
\usepackage[colorlinks,linkcolor=red,anchorcolor=blue,citecolor=red]{hyperref}

\begin{document}

\title{Learning Kernel for Conditional Moment-Matching Discrepancy-based Image Classification}

\author{Chuan-Xian~Ren, Pengfei Ge, Dao-Qing~Dai, Hong~Yan
\thanks{C.X. Ren, P.F. Ge, and D.Q. Dai are with the School of Mathematics, Sun Yat-Sen University, Guangzhou, China. C.X. Ren and P.F. Ge contribute equally to this work. Corresponding author: C.X. Ren.}
\thanks{H. Yan is with the Department of Electronic Engineering, City University of Hong Kong, Hong Kong.}
\thanks{This work is supported in part by the National Natural Science Foundation of China under Grants 61572536, 11631015, and U1611265, in part by the Science and Technology Program of Guangzhou under Grant 201804010248, and in part by the Hong Kong Research Grants Council (Project C1007-15G).}}

\date{}
\IEEEcompsoctitleabstractindextext{%
\begin{abstract}%
Conditional Maximum Mean Discrepancy (CMMD) can capture the discrepancy between conditional distributions by drawing support from nonlinear kernel functions, thus it has been successfully used for pattern classification. However, CMMD does not work well on complex distributions, especially when the kernel function fails to correctly characterize the difference between intra-class similarity and inter-class similarity. In this paper, a new kernel learning method is proposed to improve the discrimination performance of CMMD. It can be operated with deep network features iteratively and thus denoted as KLN for abbreviation. The CMMD loss and an auto-encoder (AE) are used to learn an injective function. By considering the compound kernel, i.e., the injective function with a characteristic kernel, the effectiveness of CMMD for data category description is enhanced. KLN can simultaneously learn a more expressive kernel and label prediction distribution, thus, it can be used to improve the classification performance in both supervised and semi-supervised learning scenarios. In particular, the kernel-based similarities are iteratively learned on the deep network features, and the algorithm can be implemented in an end-to-end manner. Extensive experiments are conducted on four benchmark datasets, including MNIST, SVHN, CIFAR-10 and CIFAR-100. The results indicate that KLN achieves state-of-the-art classification performance.
\end{abstract}

\begin{IEEEkeywords}
Conditional distribution discrepancy, Moment matching network, Supervised learning, Semi-supervised learning, Auto-Encoder, Kernel mappings
\end{IEEEkeywords}}
\maketitle \IEEEdisplaynotcompsoctitleabstractindextext \IEEEpeerreviewmaketitle

\section{Introduction}

\IEEEPARstart{D}{ata} classification provides us fundamental knowledge for structured information and latent connection in given data. In machine learning, data classification refers to the problem of identifying to which category a new observation belongs, under the assumption that the training set containing observations whose category membership is known or partially known. Accordingly, classification methods are usually categorized into supervised methods and semi-supervised methods. It is worth noting that practical data are usually high dimensional with large variations, which bring additional difficulty in classification. Taking the street view image as an example, data can be captured in crowds, and they have very complex backgrounds~\cite{Wang2018street,Wang2019Scene}. Thus, how to learn discriminative and compact feature, which also has good generalization performance, is a challenging and important problem.

In the past two decades, classification models are usually accompanied by exploration of nonlinear kernel functions, which enable classification models to be operated in a Reproducing Kernel Hilbert Space (RKHS)~\cite{smola2007hilbert}. Generally, RKHS corresponds to a high-dimensional, implicit but more separable feature space without explicitly computing data coordinates in that space, but rather by simply computing the inner products between all sample pairs in the feature space. It has been validated that increasing the data dimension to very high even infinite by using kernel mappings and then implementing dimension reduction operations can effectively improve the classification accuracy. It also becomes potential mechanism and important experience for data analysis and feature representation~\cite{Wang2017Discriminative,Ren2016Enhanced,Lin2013nin}. Classification algorithms capable of operating with kernels include Perceptron~\cite{Herbster07Perceptron}, Support Vector Machine~\cite{Feng2016robust}, Fisher's linear discriminant analysis~\cite{witten2011penalized,ren2015sample}, and several others~\cite{ren2014most,Yu2018MOLP}.

A probability distribution can also be embedded into RKHS by kernel mappings, and then some linear methods can be used to deal with higher-order statistics~\cite{smola2007hilbert,sriperumbudur2008injective}. The widely used methods are usually based on MMD and CMMD. MMD captures the discrepancy between marginal distributions, while CMMD captures the discrepancy between conditional distributions~\cite{fukumizu2008kernel}. These measures are widely used in independent test~\cite{fukumizu2008kernel}, non-parametric test~\cite{gretton2007kernel}, image generation~\cite{Li2015Generative,Li2017mmdgan}, and transfer learning tasks~\cite{Sun2016Return,Ren2018GCDA,Yan2017mind}.
	
MMD has been successfully applied to measure the difference between two probability distributions via kernel mean embedding. For supervised classification tasks, MMD can be used to estimate the similarity between distributions of positive and negative samples, and maximizing MMD leads to the two classes as apart as possible~\cite{Kenji2009Kernel}. It has been proved that MMD of binary-class samples in a certain condition is inverse proportional to the optimal risk of the linear loss function of Parzen window classifier~\cite{Kenji2009Kernel}. However, it is difficult to deal with multi-class classification tasks using MMD. In~\cite{Ren2016Conditional}, the authors propose conditional generative moment-matching network (CGMMN), which exploits CMMD to build conditional moment matching networks for image classification and generation tasks. Let $X$ be an observation and $Y$ be the response. CMMD measures the discrepancy between conditional distributions $P(Y|X)$ and $P(\widehat{Y}|X)$, in which $\widehat{Y}$ denotes the predicted value of $Y$. By minimizing the CMMD measurement, the multi-class classification objectives can be achieved. However, it still has the following limitations. First, CGMMN uses a fixed Gaussian kernel, which limits the expressiveness of CMMD. Second, CGMMN requires the ground-truth label for each input sample, which makes it impossible to deal with semi-supervised classification tasks.

Moreover, both MMD and CMMD cannot work well on complex distributions since it is difficult to find a suitable kernel function. Long et al.~\cite{long2015learning} propose a kernel selection strategy by maximizing MMD to find a weighted coefficient of the multi-kernel function. Li et al.~\cite{Li2017mmdgan} propose to learn the kernel function by compounding of a characteristic kernel and an injective function. The same problems appear in CMMD. When dealing with the image dataset with complex background, the kernel may not effectively characterize similarity between samples, and then the classification performance based on CMMD degrades significantly. Meanwhile, since CMMD is calculated based on the embedding of conditional distributions, the kernel learning method used in MMD cannot be directly applied in CMMD. Therefore, how to design an effective and distinguishable kernel is the most important task in the CMMD-based data classification algorithms.

In this paper, we propose a kernel learning method, i.e., KLN, to tackle these problems. KLN can simultaneously learn a more representative kernel function and label prediction distribution through CMMD. The CMMD between two similar conditional distributions is minimized to find the representative kernel function, which is compounded by a pre-specified kernel function and a deep network. In order to make the learned kernel function characteristic, an additional AE structure is used to ensure that the transformation function represented by the deep network is approximately injective. KLN is mostly proposed to deal with supervised classification tasks by minimizing the CMMD between $P(Y|X)$ and $P(\widehat{Y}|X)$, however, it can be extended to deal with semi-supervised classification tasks by using dynamically predicted labels. We evaluate KLN in a wide range of tasks, including supervised classification, visualization of the learned kernel (similarity) and semi-supervised classification. Extensive experiments are conducted on various datasets, and the results show that KLN can obtain very competitive performance.

Our contributions are summarized as follows.
\begin{enumerate}
\item It is observed that the kernel functions used in CMMD cannot effectively represent similarities between sample pairs. An inappropriate kernel function will lead to degradation of the classification performance. A kernel learning method, KLN, is proposed to tackle this problem. KLN approximates the kernel matrix in a feature space mapped by a deep network, rather than the high dimensional input features.
  \item By simultaneously learning the kernel function and the label prediction distribution, an end-to-end training algorithm is proposed to improve the final classification performance. In addition, the algorithm is extended to deal with semi-supervised classification tasks.
  \item KLN is evaluated on several benchmark image datasets. It achieves competitive prediction accuracies in both supervised and semi-supervised classification tasks. The supervised prediction accuracies on the MNIST, SVHN, CIFAR-10 and CIFAR-100 datasets achieve to 99.61\%, 98.44\%, 94.85\% and 77.37\%, respectively. This is also the first CMMD-based work to achieve competitive results in semi-supervised classification scenarios.
\end{enumerate}

The rest of this paper is organized as follows. Section~\ref{sect:related-works} briefly reviews closely related work. In Section~\ref{sect:KLN}, we first demonstrate the importance of kernel function in CMMD, and then we propose the KLN algorithm to deal with supervised and semi-supervised classification tasks. Experiment results and analysis are presented in Section~\ref{sect:experiments}, where KLN is compared with several state-of-the-art methods. Section~\ref{sect:conclusion} concludes the paper and discusses future work.

\section{Related work}\label{sect:related-works}

Generative models belong to statistical methodology of the joint probability distribution on both observation variable and response variable. Deep generative models characterize the distribution of observations with a hierarchical architecture and many latent variables. They can be the natural choice for many tasks that require statistical inference and deep convolutional operations, such as image generation~\cite{Hinz2018gener,Radford2015dcgan}, style transfer~\cite{Li2017mmdgan}, and data classification~\cite{Mirza2016CGAN,Springenberg2016CatGAN,Wang2019getnet}.

Recently, Goodfellow et al.~\cite{Goodfellow2014GAN} present generative adversarial network (GAN) to deal with the distribution learning problem. GAN adopts a game-theory-based strategy to discriminatively learn the data generator, and formulates the objective function as a min-max optimization problem~\cite{Springenberg2016CatGAN, Tzeng2017Adversarial}. However, the adversarial formalism of GAN is hard to convergent to the desired solution, as the gradients can be easily saturated and even disappear. In particular, Li et al.~\cite{Li2015Generative} present generative moment matching network (GMMN), which is essentially a generative version of MMD, to simplify the model by sampling from some simple distribution. To learn the network weights, kernel-based MMD is exploited to avoid unnecessary assumptions of the distributions. However, GMMN has some drawbacks in real applications. Besides the quadratic computational complexity, which is principally deduced by MMD, the gradient vanishing phenomenon frequently appears for low-bandwidth kernels. It is probably that some kernels used in practice are unsuitable for capturing very complex distances in high dimensional sample spaces such as natural images~\cite{Li2017mmdgan}. A GMMN network estimates the joint distribution of a set of variables. It seems that conditional distribution matching is more interesting in many other cases such as data classification and generation~\cite{Denton2015Deep, ng2002discriminative}. Ren et al.~\cite{Ren2016Conditional} present CGMMN to learn a flexible conditional distribution when some input variables are given. CGMMN extends the capability of GMMN to address a wide range of application problems as mentioned above, while keeping the training process simple. The successful applications of CGMMN can be primarily attributed to the flexible kernel embedding of conditional probability distributions, which rely on a generalization of covariance matrix, known as the cross-covariance operator~\cite{Baker1973Joint}. Song et al.~\cite{Song2009Hilbert} have proved that under some assumptions, the conditional embedding exists, and it can be expressed in terms of cross-covariance operator. Specifically, CGMMN attempts to learn the true category distribution by minimizing the CMMD between two conditional distributions.

\begin{figure*}[htb]
\subfigure[]{\label{fig:mnist_C_1}
\begin{minipage}[c]{0.5\textwidth}
\centering \scalebox{0.4}{
\includegraphics{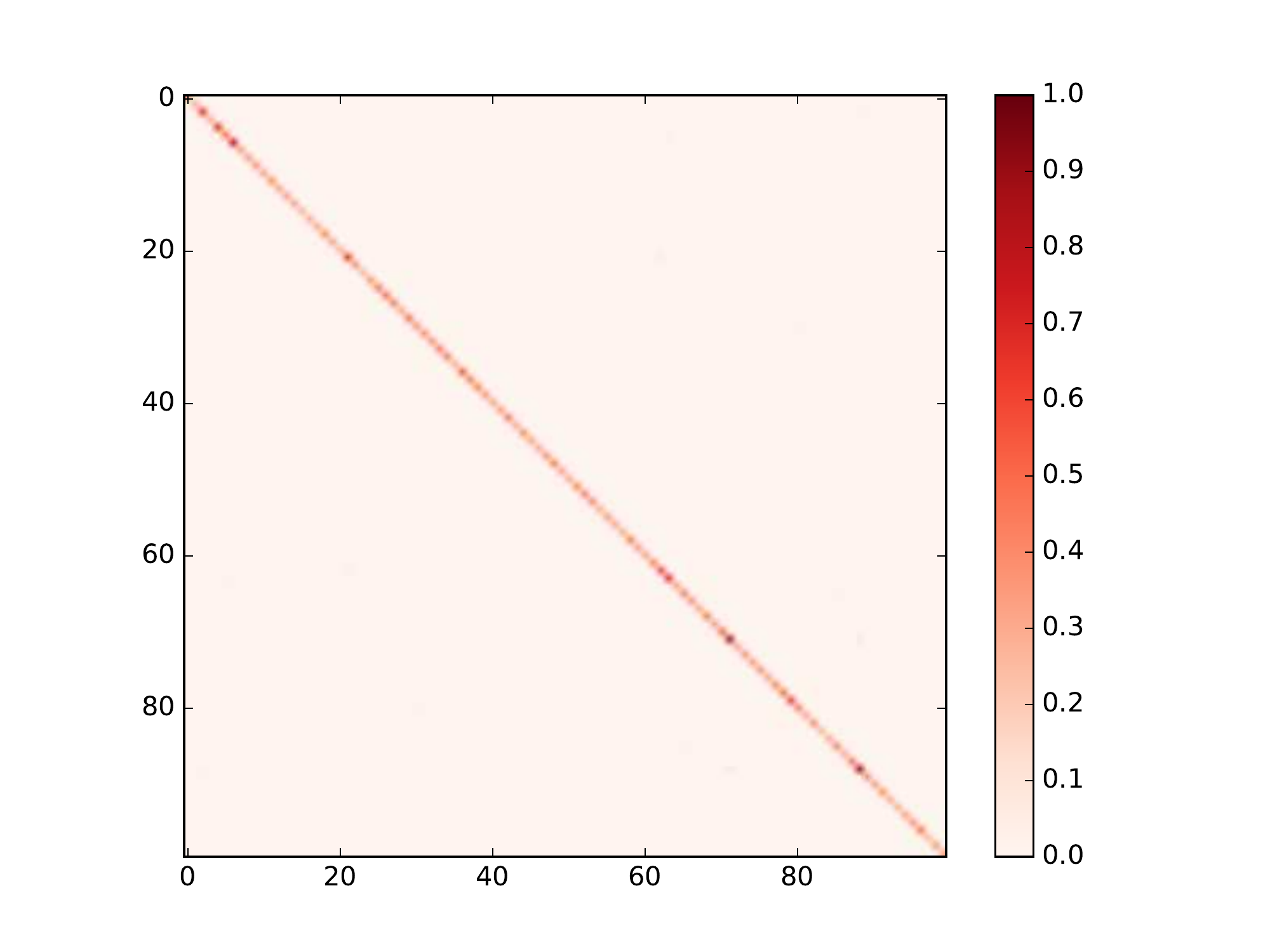}}
\end{minipage}}
\subfigure[]{\label{fig:ones_mnist_C_1}
\begin{minipage}[c]{0.5\textwidth}
\centering \scalebox{0.4}{
\includegraphics{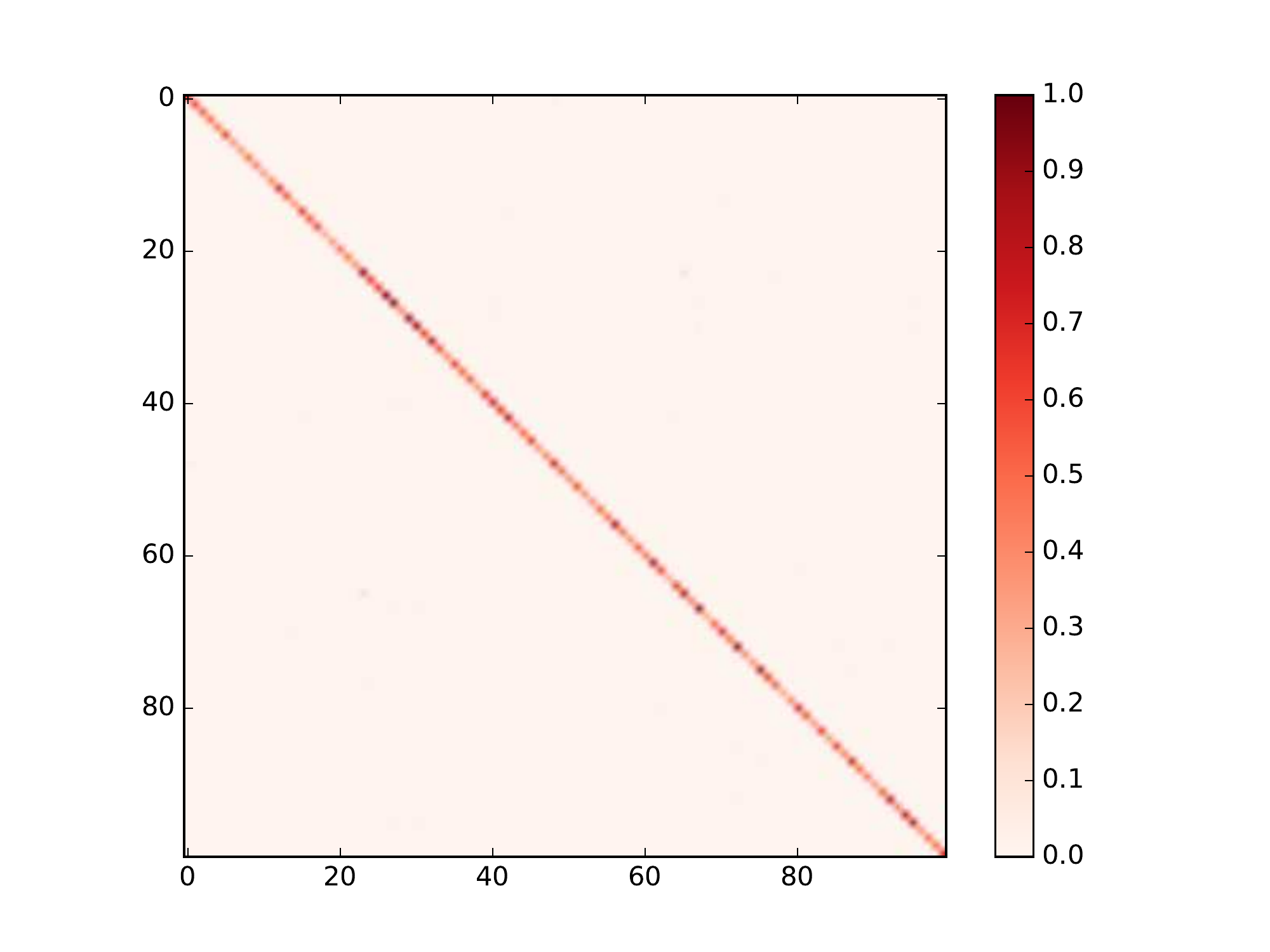}}
\end{minipage}}
\caption{Heat map of matrix $\textbf{H}$ on MNIST dataset. (a) The samples were drawn from different classes. (b) The samples were drawn from the same class. We can see that the off-diagonal elements of $\textbf{H}$ are almost identical to each other, and there is almost no difference between the two matrices. It indicates that the kernel functions used in CMMD criterion fails to distinguish between-class sample similarities and within-class sample similarities.}\label{fig:MNIST_C_1}
\end{figure*}	

The KLN method is inspired by CGMMN, but there are two key differences in the training process:
\begin{enumerate}
  \item CGMMN uses a pre-specified kernel on the raw input features directly, while KLN introduces the kernel learning process to improve the expressiveness of CMMD.
  \item CGMMN estimates CMMD on each mini-batch, which makes it necessary to have a real label for each sample. However, by simultaneously learning the kernel function and predicting the distribution by CMMD, KLN can use the unlabeled data to learn a more discriminant kernel function and estimate the CMMD between the true conditional distribution and the predicted distribution. This enables it to deal with semi-supervised classification tasks effectively.
\end{enumerate}
	
\section{The KLN Algorithm}\label{sect:KLN}

In this section, the KLN method is introduced in details to learn more expressive kernel functions and improve the classification performance. Roughly speaking, KLN learns discriminative features in an end-to-end manner by integrating CMMD and an AE framework. In particular, the kernel-based similarities are obtained by using the embedding features, rather than the input raw features. Thus, the kernel matrices are not unity in the training process and they make final features more adaptive and discriminative.

\subsection{Kernel degradation in CMMD}\label{sect:CGMMN}

The principal problem that pattern classification concerns is whether conditional distribution equation $P(Y|X)=P(\widehat{Y}|X)$ holds or not. Intuitively, CMMD is suitable for solving the pattern classification task since it can capture the discrepancy between two conditional distributions. Suppose that $\phi$ and $\psi$ denote the nonlinear mapping functions for $X$ and $Y$, respectively. Then cross-covariance $C_{XY}: \mathcal{G}\rightarrow \mathcal{F}$ is defined as~\cite{Baker1973Joint}:
\begin{equation*}
  C_{XY}=\mathbb{E}_{XY}[\phi (X) \otimes\psi (Y)]-\mu_X\otimes\mu_Y,
\end{equation*}
where  $\otimes$ is the tensor product operator. The embedding of conditional distributions $C_{Y|X}$ can be defined as~\cite{Song2009Hilbert}:
\begin{equation}\label{eq:cond-embed}
C_{Y|X}=C_{YX}C_{XX}^{-1}.
\end{equation}
Given a dataset $\mathcal{D}_{XY}$ drawn $i.i.d$ from $P(X, Y)$, $C_{Y|X}$ can be estimated by:
\begin{eqnarray}\label{eq:empir-cond-embed}
\nonumber  \widehat{C}_{Y|X} &=& \Psi(\Phi^\top\Phi+\lambda\textbf{I}) ^{-1}\Phi^\top \\
                                                        &\triangleq& \Psi(\mathcal{K}+\lambda\textbf{I}) ^{-1}\Phi^\top,
\end{eqnarray}
where $\Phi$ denotes the embedding of $X$, $\Psi$ denotes the embedding of $Y$, $\mathcal{K}=\Phi^\top\Phi$ is the Gram matrix, and $\lambda$ is a positive regularization parameter.

We randomly sample two batches, e.g., $\mathcal{D}_{XY}^s=\{(\textbf{x}_i^s, y_i^s) \}_{i=1}^n$ and $\mathcal{D}_{X\hat{Y}}^t=\{(\textbf{x}_i^t, \widehat{y}_i^t) \}_{i=1}^n$, from $P(Y|X)$ and $P(\widehat{Y}|X)$, respectively. Let $\textbf{X}_s=[\textbf{x}_1^s,\cdots, \textbf{x}_n^s]$ and $\textbf{X}_t=[\textbf{x}_1^t,\cdots, \textbf{x}_n^t]$, and the superscripts or subscripts $s$ and $t$ represent the two sample sets, respectively. $\hat{y}_i^t$ is the predicted label of $x_i^t$, and it is derived by some specified prediction method such as softmax. It is worth noting that $\textbf{X}_s$ and $\textbf{X}_t$ may be non-overlapping, because in some applications, e.g., semi-supervised classification tasks, there are some unlabeled input data. The empirical estimation of CMMD is
\begin{equation}\label{eq:empir-cmmd}
  \begin{aligned}
	L_{\rm{CMMD}}=&\|\widehat{C}_{Y|\textbf{X}}^s-\widehat{C}_{\widehat{Y}|\textbf{X}}^t\|_{\mathcal{F\otimes G}}^2\\
	=&\|\Psi_s(K_s+\lambda\textbf{I})^{-1}\Phi_s^\top-\Psi_t(K_t+\lambda \textbf{I})^{-1}\Phi_t^\top\|_{\mathcal{F\otimes G}}^2\\ =&\mbox{Tr}(K_s\widetilde{K}_s^{-1}\mathcal{L}_s\widetilde{K}_s^{-1})+\mbox{Tr}(K_t\widetilde{K}_t^{-1}\mathcal{L}_t\widetilde{K}_t^{-1})\\
&-2\cdot \mbox{Tr}(K_{ts}\widetilde{K}_s^{-1}\mathcal{L}_{st}\widetilde{K}_t^{-1}),
	\end{aligned}
\end{equation}
where $\Psi_s=[\psi (y_1^s),\cdots, \psi (y_n^s)]$,	$\Phi_s=[\phi (\textbf{x}_1^s),\cdots, \phi (\textbf{x}_n^s)]$, $K_s=\Phi_s^\top\Phi_s$, $\widetilde{K}_s=K_s+\lambda \textbf{I}$, $\mathcal{L}_s=\Psi_s^\top\Psi_s$. Other variables relating to subscribe $t$ are defined in a similar way on dataset $\mathcal{D}_{X\hat{Y}}^t$. By minimizing this CMMD-based loss function, the difference between $P(Y|X)$ and $P(\widehat{Y}|X)$ is reduced to a reasonable range. In particular, according to Theorem \textbf{3} shown in~\cite{Ren2016Conditional}, when CMMD approaches its minimum value of 0, the predicted distribution converges to the true distribution. It means that the CMMD criterion can be used to learn the category distribution.

However, CMMD fails to measure the similarity between two conditional distributions if the kernel function is selected improperly. It is well known that the performance of MMD is limited by the quality of the kernel function~\cite{Gretton2012Optimal,Zhang2013Domain}. When dealing with complex distributions, it is difficult to effectively represent similarity between samples using the RBF kernel and the Laplacian kernel, which lead to the degradation of MMD-based algorithms. The same problem exists in the CMMD criterion. We present a demonstrative instance here. We randomly select a sample batch and plot the heat map of the matrix $\tilde{K}^{-1}K\tilde{K}^{-1}$, which is the weight matrix of $\mathcal{L}$ in Equation~\eqref{eq:empir-cmmd} and is called $\textbf{H}$ here for convenience. Figure~\ref{fig:mnist_C_1} shows the $\textbf{H}$ matrix by using the samples with random sampling. The off-diagonal elements of $\textbf{H}$ are almost the same. Besides, we randomly select a sample batch from one single class, and then plot the heat map of matrix $\textbf{H}$ in Figure~\ref{fig:ones_mnist_C_1}. There is almost no difference between the two matrices. It indicates that the kernel function used in CMMD criterion fails to distinguish the between-class similarities and the within-class similarities when dealing with complex distributions. The heat maps of $\textbf{H}$ matrix on the SVHN and CIFAR-10 datasets show very similar comparison results, and they are presented in Appendix~\ref{sect:appendix-heatmap}.

In summary, kernel function learning module plays an important role in characterizing the loss function $L_{\rm{CMMD}}$. If the kernel functions are not learned well, then CMMD loses its intrinsic discriminative representation ability. In contrary, well-learned kernel functions are useful to extract robust and discriminative features for classification. Therefore, a better kernel function is expected to improve the classification performance of the CMMD criterion.		

\subsection{Kernel learning on network embedding features}

\begin{figure*}[htbp]
\centering{{\includegraphics[width=6.5in]{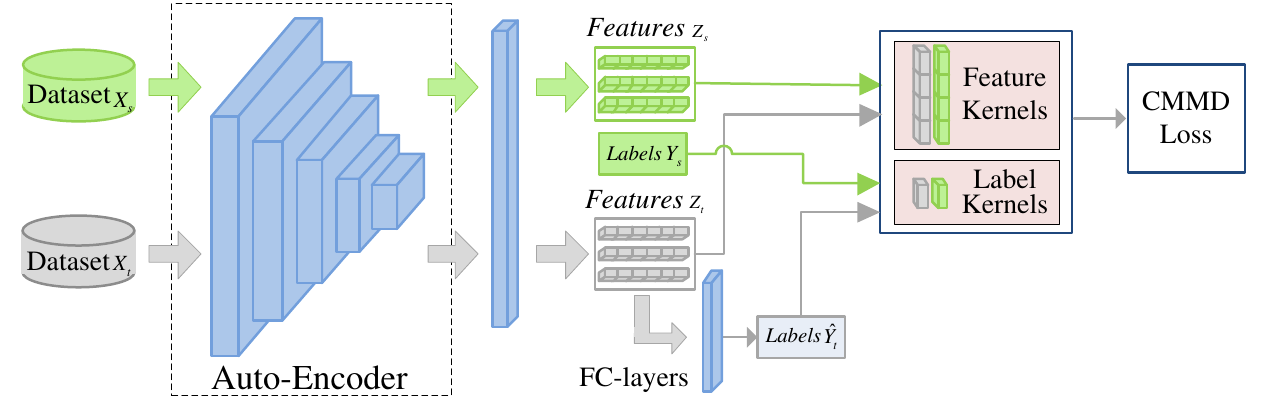}}
\caption{Flowchart of KLN method, in which the kernel matrices are computed by using the AE latent features $\textbf{Z}_s$, $\textbf{Z}_t$ and their (predicted) labels. The Auto-Encoder module is used to ensure that the mapping from $\textbf{X}$ to $\textbf{Z}$ is approximately injective. Detailed setups and parameters are placed in Appendix~\ref{sect:appendix-network}.}\label{fig:flowchart}}
\end{figure*}

In classification tasks, it is desirable that distances between samples with the same label are small in the learned metric space, while distances between those with different labels are large. However, as shown in Figure~\ref{fig:MNIST_C_1}, when the nonlinear kernel functions are directly applied to input features $\textbf{X}_s$ and $\textbf{X}_t$, the RBF kernel function tends to derive a similar Gram matrix for the high-dimensional and normalized data. In this case, CMMD cannot characterize the weights of errors well, thus the classification performance degenerates rapidly.

By Equation~\eqref{eq:empir-cmmd}, the distance $L_{\rm{CMMD}}$ of the two conditional distributions $P(Y|X)$ and $P(\tilde{Y}|X)$ can be estimated by sampling  finite samples. Due to sampling bias, $L_{\rm{CMMD}}$ may not be zero (but should be close to 0) when $P(Y|X) = P(\tilde{Y}|X)$. When $L_{\rm{CMMD}}$ is a large value, the pre-specified kernel function may be inappropriate. Intuitively, if a kernel $K$ gets a small $L_{\rm{CMMD}}$ when $P(Y|X) = P(\tilde{Y}|X)$, it is likely to describe the similarity better between samples. So, instead of directly using a pre-specified kernel $K$ for the input features $\textbf{X}$, we expect to obtain a more suitable kernel function via
\begin{equation}\label{eq:learn_k1}
  \min_{K}\|\widehat{C}_{Y|\textbf{X}}^s-\widehat{C}_{Y|\textbf{X}}^t\|_{\mathcal{F\otimes G}}^2.
\end{equation}
	
In Equation~\eqref{eq:learn_k1}, all characteristic kernels should be considered. However, it is difficult to choose the optimal solution from all the characterized kernels. According to the theory proposed by \cite{Gretton2012Optimal}, if function $h$ is injective and the kernel function $K$ is characteristic, then the compound kernel function $\mathcal{K}=K\circ h$, which satisfies to $\mathcal{K}(x,x^{'})=K(h(x),h(x^{'}))$, is still characteristic. In this paper, the optimal solution of Equation~\eqref{eq:learn_k1} is determined via selecting from a series of kernel functions $\mathcal{K}$. In other words, the kernel function is not applied directly to the raw data $X$, but instead applied to the embedded feature $Z$ with an $h$ transformation, which is represented by a deep network. Here an injective function $h_\textbf{w}$ with parameter $\textbf{w}$ is considered. Let $\textbf{Z} = h_\textbf{w}(\textbf{X})$. We use subscript $\_z$ to denote the features referring to $\textbf{Z}$ and to distinguish them from raw features $\textbf{X}$. Then the kernel learning loss by the CMMD criterion in the new model can be defined as:
\begin{equation}\label{eq:kernel}
	\begin{aligned}
	& L_{\rm{CMMD\_Z}}(\textbf{w})\\
	=&\|\widehat{C}_{Y_s|h_\textbf{w}(\textbf{X}_s)}-\widehat{C}_{Y_t|h_\textbf{w}(\textbf{X}_t)}\| _{\mathcal{F\otimes G}}^2\\
	=&\|\widehat{C}_{Y_s|\textbf{Z}_s}-\widehat{C}_{Y_t|\textbf{Z}_t}\|_{\mathcal{F\otimes G}}^2\\		=&\|\Psi_s(K_{s\_z}+\lambda\textbf{I})^{-1}\Phi_{s\_z}^\top-\Psi_t(K_{t\_z}+\lambda\textbf{I})^{-1}\Phi_{t\_z}^\top\|_{\mathcal{F\otimes G}}^2\\ =&\mbox{Tr}(K_{s\_z}\widetilde{K}_{s\_z}^{-1}\mathcal{L}_s\widetilde{K}_{s\_z}^{-1})+\mbox{Tr}(K_{t\_z}\widetilde{K}_{t\_z}^{-1}\mathcal{L}_t\widetilde{K}_{t\_z}^{-1})\\
-&2\cdot \mbox{Tr}(K_{ts\_z}\widetilde{K}_{s\_z}^{-1}\mathcal{L}_{ts}\widetilde{K}_{t\_z}^{-1}),
	\end{aligned}
\end{equation}
in which $\Psi_s=[\psi(y_1^s),\cdots,\psi(y_n^s)]$, $\mathcal{L}_s=\Psi_s^\top\Psi_s$, $\Phi_{s\_z}=[\phi(\textbf{z}_1^s),\cdots,\phi(\textbf{z}_n^s)]=[\phi(h_\textbf{w}(\textbf{x}_1^s)),\cdots,\phi(h_\textbf{w}(\textbf{x}_n^s))]$, $K_{s\_z}=\Phi_{s\_z}^\top\Phi_{s\_z}$, and $\widetilde{K}_{s\_z}\!\!=\!\!K_{s\_z}\!+\!\lambda\textbf{I}$. Other variables relating to subscribe $t$ are defined in a similar way on $\mathcal{D}_{XY}^t$. $K_{ts\_z}=\Phi_{t\_z}^\top\Phi_{s\_z}$, $\mathcal{L}_{st}=\Psi_s^\top\Psi_t$. By minimizing $L_{\rm{CMMD\_Z}}(\textbf{w})$, the similarity value of positive pair (i.e., in the same class) becomes larger, while the similarity value of negative pair (i.e., in different classes) becomes smaller.

In addition, to ensure that $h_{\textbf{w}}$ is injective or approximately injective, an auto-encoder (AE) module is designed in KLN. For an injective function $h$, there exists an inverse function $h^{-1}$ such that $h^{-1}(h(\textbf{x}) ) = \textbf{x}$, $\forall \textbf{x} \in \mathcal{X}$, which can be approximated by an AE. In the following, we use $\textbf{w}=\{\textbf{w}_e,\textbf{w}_d\}$ to denote the network parameters of AE, in which $\textbf{w}_e$ denotes the encoder parameters and $\textbf{w}_d$ the decoder parameters. Correspondingly, $h_{\textbf{w}_e}$ is the encoder and $h_{\textbf{w}_d}$ is the decoder. The reconstruction loss of AE can be defined as:
\begin{equation}\label{eq:ae}
\begin{aligned}
	L_{ae}(\textbf{w})=\mathbb{E}_{\textbf{x}\in \mathcal{X}}\|\textbf{x}-h_{\textbf{w}_d}(h_{\textbf{w}_e}(\textbf{x}))\|^2.
\end{aligned}
\end{equation}
In this AE architecture, decoder satisfies to $h_{\textbf{w}_d}\approx h_{\textbf{w}_e}^{-1}$ when the reconstruction loss is minimized, which ensures that $h_{\textbf{w}_e}$ is approximately injective.
	
Taking the AE regularization module into account, the objective function of KLN is formulated as
\begin{equation}\label{eq:kernel_learning}
\begin{aligned}
	\underset{\textbf{w}}\min\:\: L_{\rm{CMMD\_Z}}(\textbf{w}_e)+\beta L_{ae}(\textbf{w}).
\end{aligned}
\end{equation}
The first term is the CMMD criterion based on the new kernel $\mathcal{K}=K\circ h_{w_e}$, and the second term is the expected error from the AE, which can be viewed as a regularization term to ensure that the learned kernel function is characteristic. The trade-off positive parameter $\beta$ is predefined by users.

\subsection{Kernel learning for supervised classification tasks}\label{sect:super}

\begin{algorithm}[htbp]
		\caption{KLN for Supervised Classification Tasks}
		\label{Alg2}
		\begin{algorithmic}[1]
            \REQUIRE{Dataset $\mathcal{D}=\{(\textbf{x}_i,y_i)\}_{i=1}^n$}
            \ENSURE{Learned parameters $\textbf{w}=\{\textbf{w}_e,\textbf{w}_d\}, \textbf{w}_{fc}$}
			\STATE{Copy $\mathcal{D}$ into subsets $\mathcal{D}_s$ and $\mathcal{D}_t$.}
			\WHILE{Stopping criterion not met}
			\STATE{Draw a mini-batch from $\mathcal{D}_s$ and $\mathcal{D}_t$, and denote them as $\mathcal{B}_s$ and $\mathcal{B}_t$, respectively;}
			\STATE{For each $\textbf{x}\in\mathcal{B}_t$, generate $\hat{y}$; and set $\mathcal{B}^{'}_t$} to contain all the generated $(\textbf{x},\hat{y})$;
			\STATE{Solve the gradient of CMMD loss w.r.t. $\textbf{w}_e,\textbf{w}_{fc}$ on $\mathcal{B}_s$ and $\mathcal{B}^{'}_t$;}
            \STATE{Solve the gradient of AE loss w.r.t. $\textbf{w}$ on $\mathcal{B}_s$ and $\mathcal{B}^{'}_t$;}
			\STATE{Update $\textbf{w},\textbf{w}_{fc}$ using the gradient of Eq.~\eqref{eq:final-super};}
			\ENDWHILE
		\end{algorithmic}
\end{algorithm}

In the previous section, KLN is proposed based on the CMMD criterion and the AE framework. In order to deal with supervised classification tasks, a classification layer can be attached onto top of the hidden layer. As shown in Figure~\ref{fig:flowchart}, a fully connected layer with softmax activation is connected to the hidden layer. The label prediction distribution and the kernel function are learned simultaneously. In particular, the CMMD criterion is adopted to measure prediction errors. The CMMD between the prediction distribution $P(\widehat{Y}|X)$ and the true distribution $P(Y|X)$ can be estimated as:
\begin{equation}\label{eq:Supervised}
	\begin{aligned}
	L_{\rm{CMMD\_Z}}(\widehat{Y})
	=\|\widehat{C}_{Y_s|h_{\textbf{w}_e}(\textbf{X}_s)}-\widehat{C}_{\widehat{Y}_t|h_{\textbf{w}_e}(\textbf{X}_t)}\| _{\mathcal{F\otimes G}}^2.
	\end{aligned}
\end{equation}
Comparing Equations~\eqref{eq:kernel} and~\eqref{eq:Supervised}, $L_{\rm{CMMD\_Z}}(\widehat{Y})$ is an effective approximation of $L_{\rm{CMMD\_Z}}( \textbf{w}_e)$ when $P(\widehat{Y}|X) \approx P(Y|X)$. In order to reduce the computational complexity, $L_{\rm{CMMD \_Z}}(\widehat{Y})$, instead of $L_{\rm{CMMD\_Z}}(\textbf{w}_e)$, is used to learn the kernel function. In this case, the CMMD distance is calculated only once. Let $\textbf{w}_{fc}$ represent the network parameters of the fully connected layer, the objective of KLN for supervised classification tasks is modified as follows,
\begin{equation}\label{eq:final-super}
\begin{aligned}
	\underset{\textbf{w},\textbf{w}_{fc}}\min\:\: L_{\rm{CMMD\_Z}}(\widehat{Y}) + \beta L_{ae}(\textbf{w}).
\end{aligned}
\end{equation}
We call the first term as CMMD loss, and the next term as AE regularization term. In Equation~\eqref{eq:final-super}, the CMMD loss is used to learn the label prediction distribution and the kernel function simultaneously, and the AE regularization term is used to ensure that the learned kernel function is characteristic.

\begin{figure*}[htb]
\subfigure[MNIST]{\label{ones_mnist}
\begin{minipage}[c]{0.23\textwidth}
\centering \scalebox{0.35}{
\includegraphics{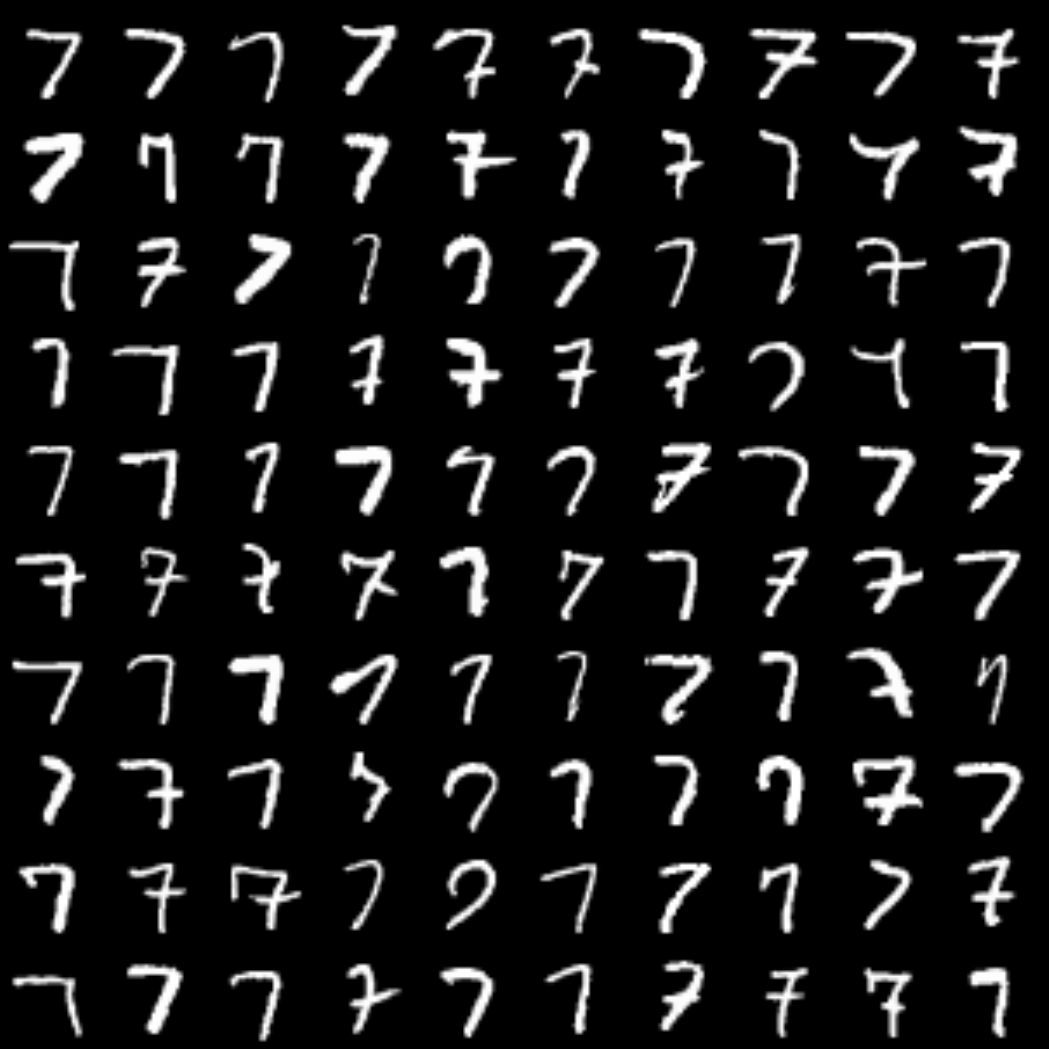}}
\end{minipage}}
\subfigure[SVHN]{\label{ones_svhn}
\begin{minipage}[c]{0.23\textwidth}
\centering \scalebox{0.31}{
\includegraphics{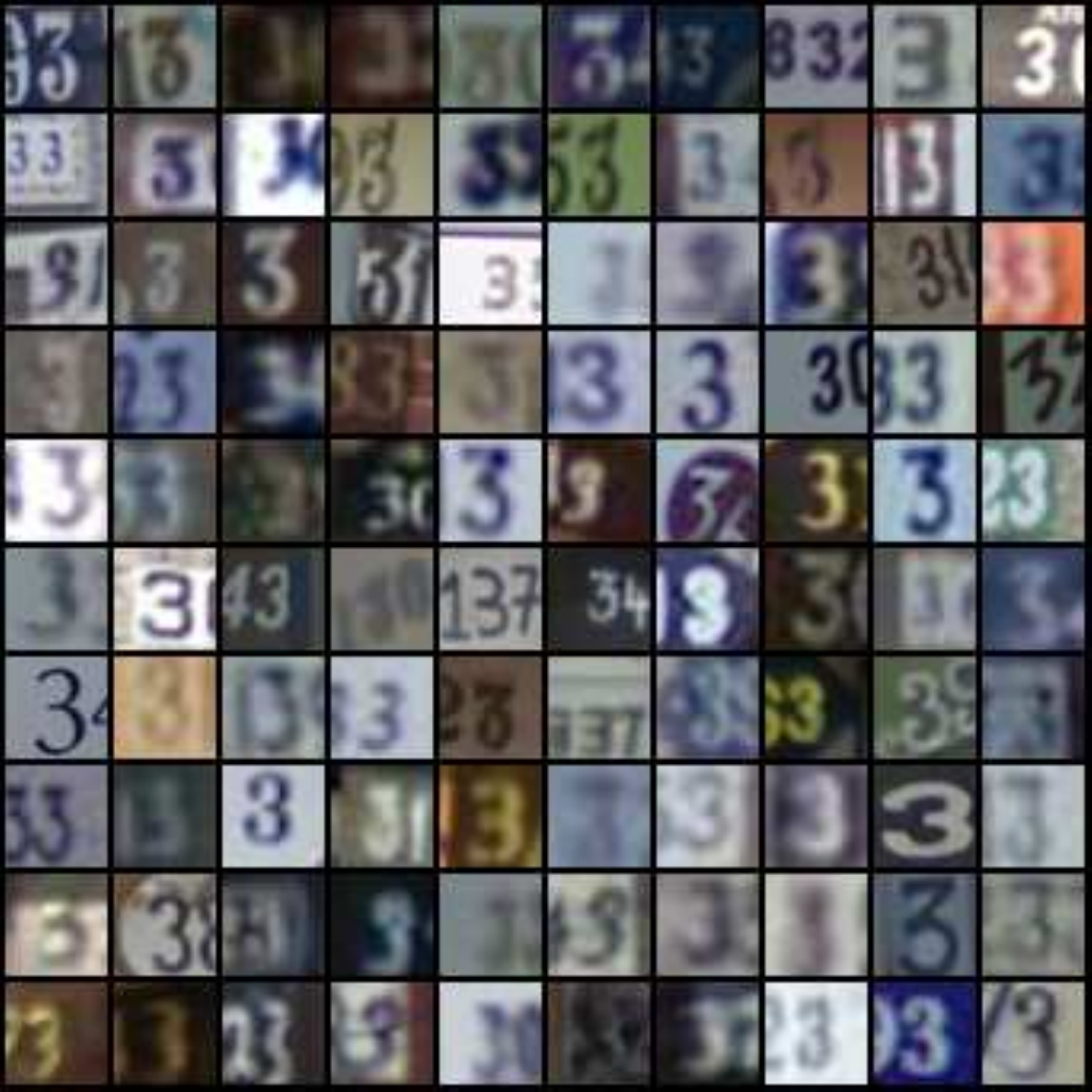}}
\end{minipage}}
\subfigure[CIFAR-10]{\label{ones_cifar10}
\begin{minipage}[c]{0.23\textwidth}
\centering \scalebox{0.39}{
\includegraphics{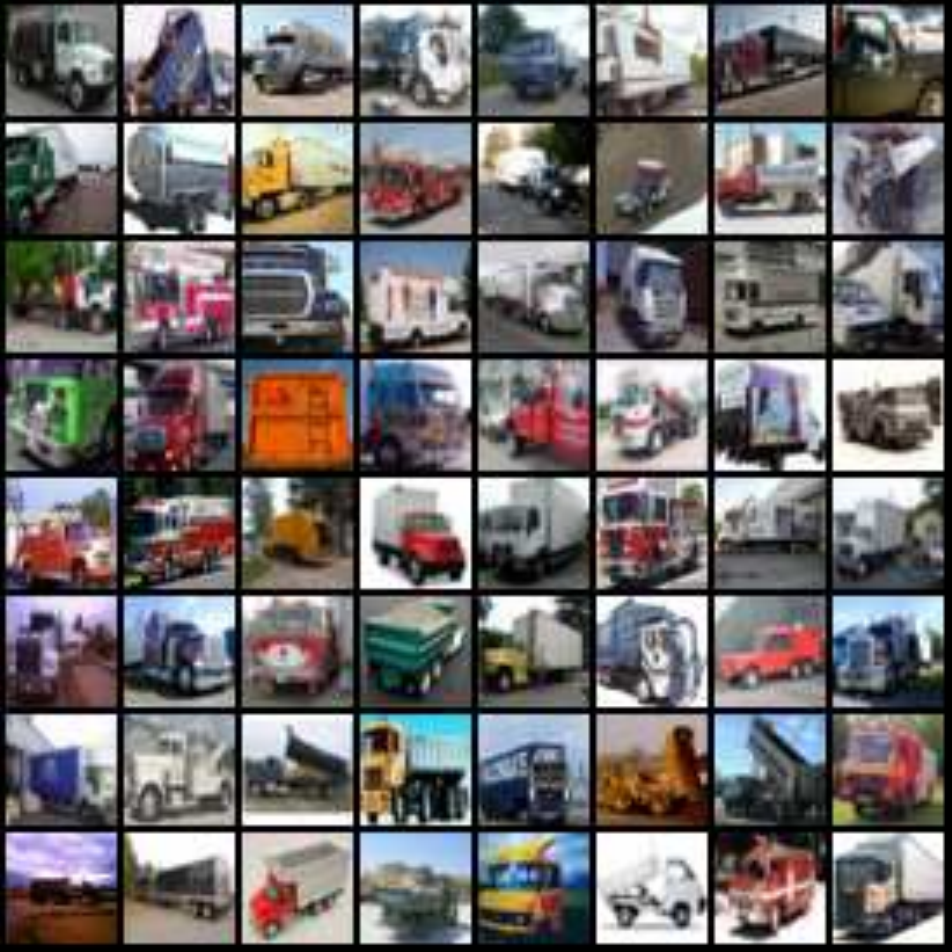}}
\end{minipage}}
\subfigure[CIFAR-100]{\label{ones_cifar10}
\begin{minipage}[c]{0.23\textwidth}
\centering \scalebox{0.31}{
\includegraphics{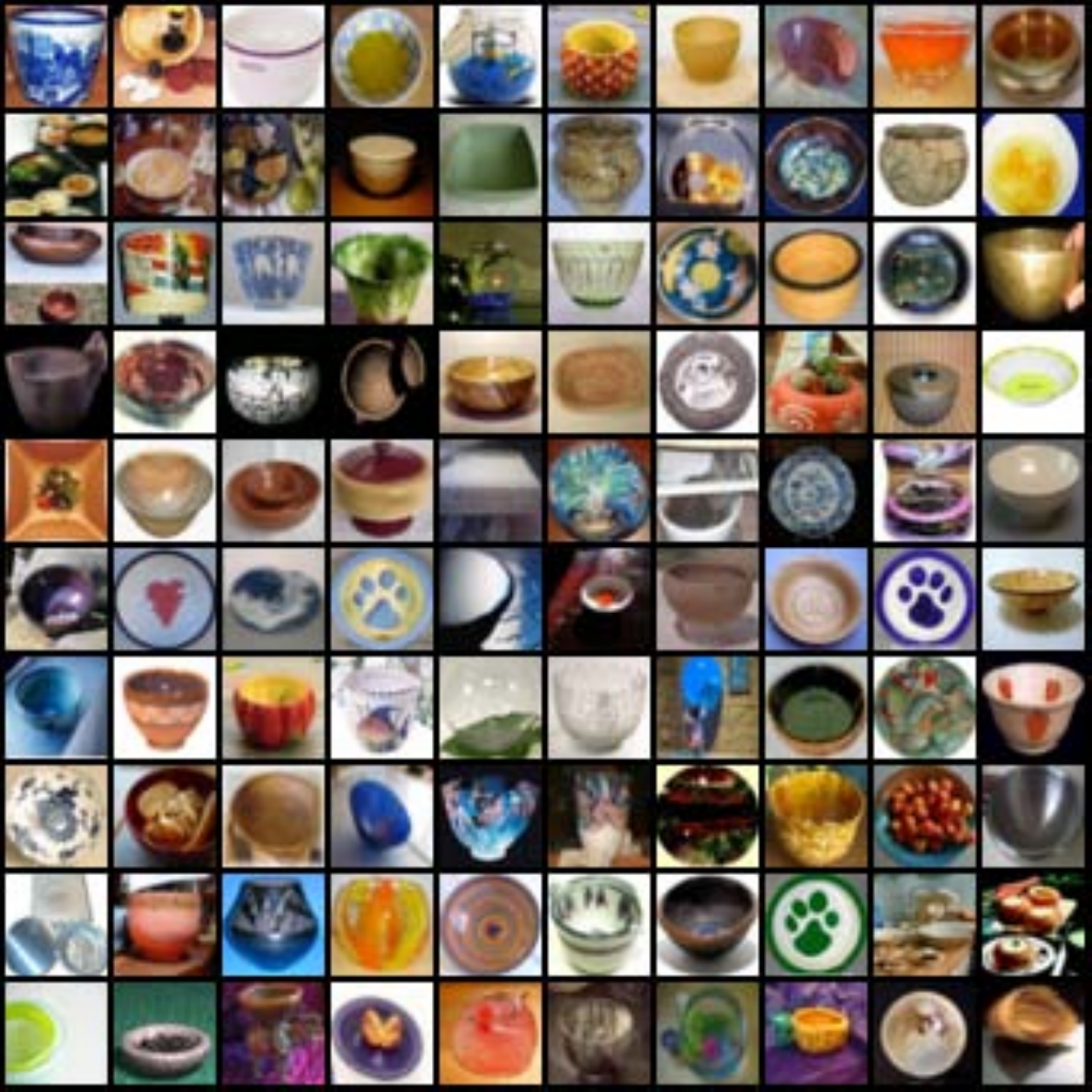}}
\end{minipage}}
\caption{Picture exemplars used in the experiments.}\label{fig:samples}
\end{figure*}

Figure~\ref{fig:flowchart} shows the flowchart of our KLN method. KLN uses the encoder to map the high-dimensional raw data $\textbf{X}$ to the latent feature $\textbf{Z}$, which can be passed through the decoder to reconstruct $\textbf{X}$. To predict label $Y$, the latent variable $\textbf{Z}$ is propagated to the loss layer via a non-linear transformation with FC-layers. Main steps of KLN for supervised classification tasks are summarized and then shown in Algorithm \ref{Alg2}.

When the training process shown in Algorithm~\ref{Alg2} is completed, we can obtain the prediction distribution $P(\hat{Y}|X)$, which is represented by the encoder of AE and the fully connected layer. When testing a new sample $\textbf{x}_{test}$, it is first input to the encoder to get the latent feature $\textbf{z}_{test}$, which is then input to the fully connected layer to get the predicted label $\hat{\textbf{y}}_{test}$.

\subsection{Kernel learning for semi-supervised classification tasks}

We now consider adapting the formulation from Section \ref{sect:super} to the semi-supervised setting. Let $\textbf{X}=[\textbf{x}_1,\cdots, \textbf{x}_N]$ be a set of all $N$ examples and $\textbf{X}^l=[\textbf{x}_1^l,\cdots, \textbf{x}_L^l]$ be a set of $L$ labeled examples. In particular, $\textbf{X}^l \subset \textbf{X}$. For every $\textbf{x}_i \in \textbf{X}^l$, the ground-truth label $y_i \in {1,\cdots,C}$ is known, where $C$ is the number of classes.

In Equation~\eqref{eq:Supervised}, $\textbf{X}_s$ needs its true label and $\textbf{X}_t$ can use the predicted label, so $\textbf{X}_s$ can be sampled from $\textbf{X}^l$ and $\textbf{X}_t$ from $\textbf{X}$ to estimate the CMMD loss. It means that in semi-supervised classification tasks, the CMMD loss can simultaneously use a small amount of labeled data and a large amount of unlabeled data to learn a more suitable kernel function. In addition, to make better use of the unlabeled data, we also introduce the confidence loss~\cite{Springenberg2016CatGAN} for these data, which is commonly used in the semi-supervised learning models. The confidence loss $L_{con}$ minimizes the conditional entropy of $P(\widehat{Y}|\textbf{X})$ and the cross entropy between $P(Y)$ and $P(\widehat{Y})$. Accordingly, the objective of KLN for semi-supervised classification tasks is presented as follows,
\begin{equation}\label{eq:final-semi}
\begin{aligned}
	\underset{\textbf{w},\textbf{w}_{fc}}\min\:\: L_{\rm{CMMD\_Z}}(\widehat{Y}) + \beta_{1} L_{ae}(\textbf{w}) + \beta_{2} L_{con}(\textbf{w}_{e},\textbf{w}_{fc}),
\end{aligned}
\end{equation}
where $\beta_{1}$ and $\beta_{2}$ are trade-off parameters. Comparing with the training process for supervised tasks, the optimization process and the sampling process are different. Main steps of KLN for semi-supervised classification tasks are summarized and then shown in Algorithm \ref{Alg3}.

\begin{algorithm}
		\caption{KLN for Semi-Supervised Classification Tasks}
		\label{Alg3}
		\begin{algorithmic}[1]
            \REQUIRE{Datasets $\mathcal{D}=\{(\textbf{x}_i)\}_{i=1}^N$, $\mathcal{D}^l=\{(\textbf{x}_i^l,y_i^l)\}_{i=1}^L$}
            \ENSURE{Learned parameters $\textbf{w}=\{\textbf{w}_e,\textbf{w}_d\}, \textbf{w}_{fc}$}
			\WHILE{Stopping criterion not met}
			\STATE{Draw a mini-batch from $\mathcal{D}^l$ and $\mathcal{D}$, and denote them as $\mathcal{B}_s$ and $\mathcal{B}_t$, respectively;}
			\STATE{For each $\textbf{x}\in\mathcal{B}_t$, generate $\hat{y}$; and set $\mathcal{B}^{'}_t$} to contain all the generated $(\textbf{x},\hat{y})$;
			\STATE{Solve the gradients of CMMD loss and AE regularization term w.r.t. $\textbf{w},\textbf{w}_{fc}$ on $\mathcal{B}_s$ and $\mathcal{B}^{'}_t$;}
            \STATE{Solve the gradient of Confidence loss w.r.t. $\textbf{w}_e,\textbf{w}_{fc}$ on $\mathcal{B}^{'}_t$;}
			\STATE{Update $\textbf{w},\textbf{w}_{fc}$ using the gradient of Eq.~\eqref{eq:final-semi};}
			\ENDWHILE
		\end{algorithmic}
\end{algorithm}
	
KLN learns discriminative features in an end-to-end manner. The loss functions~\eqref{eq:final-super} and \eqref{eq:final-semi} can be optimized by the Back-Propagation algorithm, which can be implemented automatically and efficiently by several deep learning frameworks including TensorFlow and PyTorch. Thus, the derivative details are omitted here. More implementation details will be discussed in the next section.

\section{Experiment results and analysis}\label{sect:experiments}

In this section, classification performance of KLN was evaluated from several views. The accuracy was firstly tested and compared with other methods in supervised classification tasks. Then several ablation experiments were conducted to show the importance of kernel learning. The histograms of kernel values learned by KLN are also shown. Finally, the classification results of KLN in semi-supervised classification tasks were tested and compared with other methods.

\subsection{Datasets and experimental setup}

Four benchmark datasets, i.e., MNIST~\cite{Lecun1998Gradient}, SVHN~\cite{netzer2011svhn}, CIFAR-10 and CIFAR-100~\cite{cifar-10}, were used for evaluating the image classification performance in this section. Some exemplar instances are shown in Figure~\ref{fig:samples}.

MNIST handwritten digits set is one of the most popular datasets used in the deep learning literature. It has 70,000 images of ten classes (from 0 to 9). Each sample is a 28$\times$28 gray image. The whole dataset was divided into three non-overlapping parts: one training set of 50,000 samples, one validation set of 10,000 samples, and one test set of 10,000 samples.

SVHN is a more complex dataset than MNIST as it has more samples and is closer to real scenes. Each sample is a $3\times32\times32$ color image. All the images were captured from house numbers in Google street views. The whole set is also composed of three different subsets, namely, one training set consisting of 73,257 samples, one testing set consisting of 26,032 samples, and one extra training set consisting of 531,131 samples.

CIFAR-10 is another image dataset. It consists of ten-class images including airplanes, cars, cats, and so on. Each class has 6,000 color images of size $3\times32\times32$. The whole set was divided into two separate parts, i.e., one training set of 50,000 images, and one test set of 10,000 images.

CIFAR-100 contains 60,000 color images of size $3\times32\times32$ drawn from 100 classes. The dataset was split into 50,000 training images and 10,000 test images.

All experiments were implemented by PyTorch Toolbox, and ran on a PC equipped with a NVIDIA TiTan X GPU and 32G RAM.
		
\textbf{Data processing:} For MNIST, no preprocessing was applied to the data except for scaling to the numerical range $[0,1]$. SVHN and CIFAR datasets require more techniques for image preprocessing. In addition to the global normalization, a series of data augmentation methods including random cropping and horizontal flipping were applied.
	
\textbf{Network architecture:} On the MNIST dataset, we followed the architecture of DCGAN \cite{Radford2015dcgan} to design the network architecture of KLN. We replaced the pooling layers with strided convolutions, and used batch-wise normalization and LeakyReLU activation in the auto-encoder. On the SVHN, CIFAR-10 and CIFAR-100 datasets, we followed the architecture in \cite{laine2016temporal}. More specific parameter settings are placed in Appendix~\ref{sect:appendix-network}.
	
\textbf{Hyper-parameters:} For model optimization, the batch size was set to 100 for all methods. For supervised classification tasks, SGD algorithm with initial learning rate 0.02, momentum 0.9 and weight decay 0.0005 was used. On CIFAR and MNIST, the learning rate reduces by 0.2 at 50, 100, 130 epochs, and all the networks were trained by a total of 150 epochs. On SVHN, the learning rate reduced by 0.2 at 10, 20, 30 epochs, and the training procedure finished in 40 epoches. For semi-supervised classification tasks, ADAM algorithm~\cite{Kingma2014Adam} with learning rate 1$e$-3 and the two moment terms $\gamma_1=0.9$, $\gamma_2=0.99$ was used. The dimensionality $d$ of the latent variable $\textbf{Z}$ is $d=128$. The CMMD loss was calculated by using multiple Gaussian kernel functions with bandwidth parameters [1, 3, 5, 7, 9].  For the weight hyper-parameters, we simply set $\beta=0.1$ in supervised classification tasks and $\beta_1=0.1,\beta_2=1$ in semi-supervised classification tasks.

\subsection{Experiment results of supervised classification}

This section compares KLN with some state-of-the-art supervised classification methods, which include VA+Pegasos~\cite{Li2015Max}, MMVA~\cite{Li2015Max}, Stochastic Pooling~\cite{Zeiler2013Stochastic}, Network in Network (NIN)~\cite{Lin2013nin}, Maxout Network~\cite{Goodfellow2013Maxout}, DSN~\cite{Lee2015dsn}, ResNet~\cite{He2016Deep} and so on. For a fair comparison, the same network architecture was used to evaluate the performance of the most recent method CGMMN-CNN. For the rest methods, the best results are from the original work.

Table~\ref{tab:results-mnist} shows the classification results of MNIST. KLN was trained with a total of 150 epochs. The encoder and decoder of AE used a 4-layer fully convolutional network and a 4-layer de-convolutional network, respectively. The results show that KLN is competitive with various state-of-the-art competitors, e.g., CMMVA and DSN. DSN obtains the minimum error rate as it benefits from using more supervised information in every hidden layer, and KLN gets the same classification performance as DSN. In particular, we pay our main attention on comparing the results of CGMMN and KLN. Two types of architectures of CGMMN, as stated in~\cite{Ren2016Conditional}, i.e., CGMMN-MLP and CGMMN-CNN, are evaluated and then analyzed here. The results of CGMMN-MLP and CGMMN-CNN are 0.97\% and 0.45\%, respectively, while the accuracy of KLN is 0.39\%. Thus, KLN outperforms CGMMN-MLP and CGMMN-CNN.

\begin{table}[htb]
\centering
\renewcommand{\tabcolsep}{2pc} 
\renewcommand{\arraystretch}{1.2} 
\caption{Comparison with state-of-the-art methods on MNIST.}\label{tab:results-mnist}
\begin{tabular}{cc}
\hline
Method      &  Error rate (\%)\\
\hline
VA+Pegasos~\cite{Li2015Max}       	&  1.04\\
MMVA~\cite{Li2015Max}      	& 0.90\\
CVA + Pegasos~\cite{Li2015Max}       	& 1.35 \\
Stochastic Pooling~\cite{Zeiler2013Stochastic}       	& 0.47\\
NIN~\cite{Lin2013nin}     	& 0.47\\
Maxout Network~\cite{Goodfellow2013Maxout}      	& 0.45\\
CMMVA~\cite{Li2015Max}       	& 0.45\\
DSN~\cite{Lee2015dsn}      	& 0.39\\
CGMMN-MLP~\cite{Ren2016Conditional}       	& 0.97\\
CGMMN-CNN~\cite{Ren2016Conditional}       	& 0.45\\
KLN       	& \textbf{0.39}\\
\hline
\end{tabular}
\end{table}

Table~\ref{tab:results-svhn} shows the classification results of SVHN. KLN was trained with a total of 40 epochs. The encoder and decoder of AE used a 9-layer convolutional network and a 4-layer de-convolutional network, respectively. The results of CNN and CMMVA are 4.9\% and 3.09\%, respectively. DSN still obtains a lower error rate as it benefits from using more supervised information in every hidden layer. The results of DSN, CGMMN-CNN and KLN are 1.92\%, 2.01\% and 1.56\%, respectively. Thus, KLN outperforms both DSN and CGMMN, and the effectiveness of algorithmic improvement is validated again.

\begin{table}[htb]
\centering
\renewcommand{\tabcolsep}{2pc} 
\renewcommand{\arraystretch}{1.2} 
\caption{Comparison with state-of-the-art methods on SVHN.}\label{tab:results-svhn}
\begin{tabular}{cc}
\hline
Method      &  Error rate (\%)\\
\hline
CVA+Pegasos~\cite{Li2015Max}      	&  25.3\\
CNN~\cite{sermanet2012cnn}    &  4.90  \\
CMMVA~\cite{Li2015Max}       	& 3.09\\
Stochastic Pooling~\cite{Zeiler2013Stochastic}      	& 2.80\\
NIN~\cite{Lin2013nin}      	& 2.47\\
Maxout Network~\cite{Goodfellow2013Maxout}      	& 2.35\\
DSN~\cite{Lee2015dsn}       	& 1.92\\
CGMMN-CNN~\cite{Ren2016Conditional}      	& 2.01\\
KLN       	& \textbf{1.56}\\
\hline
\end{tabular}
\end{table}

Table~\ref{tab:results-cifar10} shows the classification results of CIFAR-10. KLN was trained with a total of 150 epochs. The encoder and decoder of AE used a 9-layer convolutional network and a 4-layer de-convolutional network, respectively. All the error rates of CNN, NIN, Maxout Network, and DSN are larger than 7.0\%. The results of FitNet4-LSUV, ResNet-110 and CGMMN-CNN are 6.06\%, 6.43\% and 6.61\%, respectively, while that of KLN is 5.15\%. Therefore, KLN outperforms both DSN and CGMMN-CNN on the CIFAR-10 dataset, and the effectiveness of algorithmic improvement is validated again.
\begin{table}[htb]
\centering
\renewcommand{\tabcolsep}{2pc} 
\renewcommand{\arraystretch}{1.2} 
\caption{Comparison with state-of-the-art methods on CIFAR10.}\label{tab:results-cifar10}
\begin{tabular}{cc}
\hline
Method      &  Error rate (\%)\\
\hline
ALL-CNN~\cite{springenberg2014striving}    &  7.25  \\
NIN~\cite{Lin2013nin}  & 8.81\\
Maxout Network~\cite{Goodfellow2013Maxout}      	& 9.38\\
DSN~\cite{Lee2015dsn}      	& 7.97\\
FitNet4-LSUV~\cite{Mishkin2015All}      	&  6.06\\
ResNet-110~\cite{He2016Deep}               & 6.43\\
CGMMN-CNN~\cite{Ren2016Conditional}      	& 6.61\\
KLN       	& \textbf{5.15}\\
\hline
\end{tabular}
\end{table}

Table~\ref{tab:results-cifar100} shows the results of CIFAR-100. KLN was trained with the same parameter settings as those of CIFAR-10. All the error rates of CNN, NIN, Maxout Network, and DSN are larger than 30.0\%. The results of FitNet4-LSUV, ResNet-110 and CGMMN-CNN are 27.66\%, 25.16\% and 24.84\%, respectively, while that of KLN is 22.63\%. Compared with the results of CGMMN method, KLN has more than two percent increments.
\begin{table}[htb]
\centering
\renewcommand{\tabcolsep}{2pc} 
\renewcommand{\arraystretch}{1.2} 
\caption{Comparison with state-of-the-art methods on CIFAR100.}\label{tab:results-cifar100}
\begin{tabular}{cc}
\hline
Method      &  Error rate (\%)\\
\hline
ALL-CNN~\cite{springenberg2014striving}    &  33.71  \\
NIN~\cite{Lin2013nin}  & 35.68\\
Maxout Network~\cite{Goodfellow2013Maxout}      & 34.54\\
DSN~\cite{Lee2015dsn}      	& 34.57\\
FitNet4-LSUV~\cite{Mishkin2015All}      	&  27.66\\
ResNet-110~\cite{He2016Deep}               & 25.16\\
CGMMN-CNN~\cite{Ren2016Conditional}      	& 24.84\\
KLN       	& \textbf{22.63}\\
\hline
\end{tabular}
\end{table}

\begin{figure}[h]
\centering{{\includegraphics[width=2.6in]{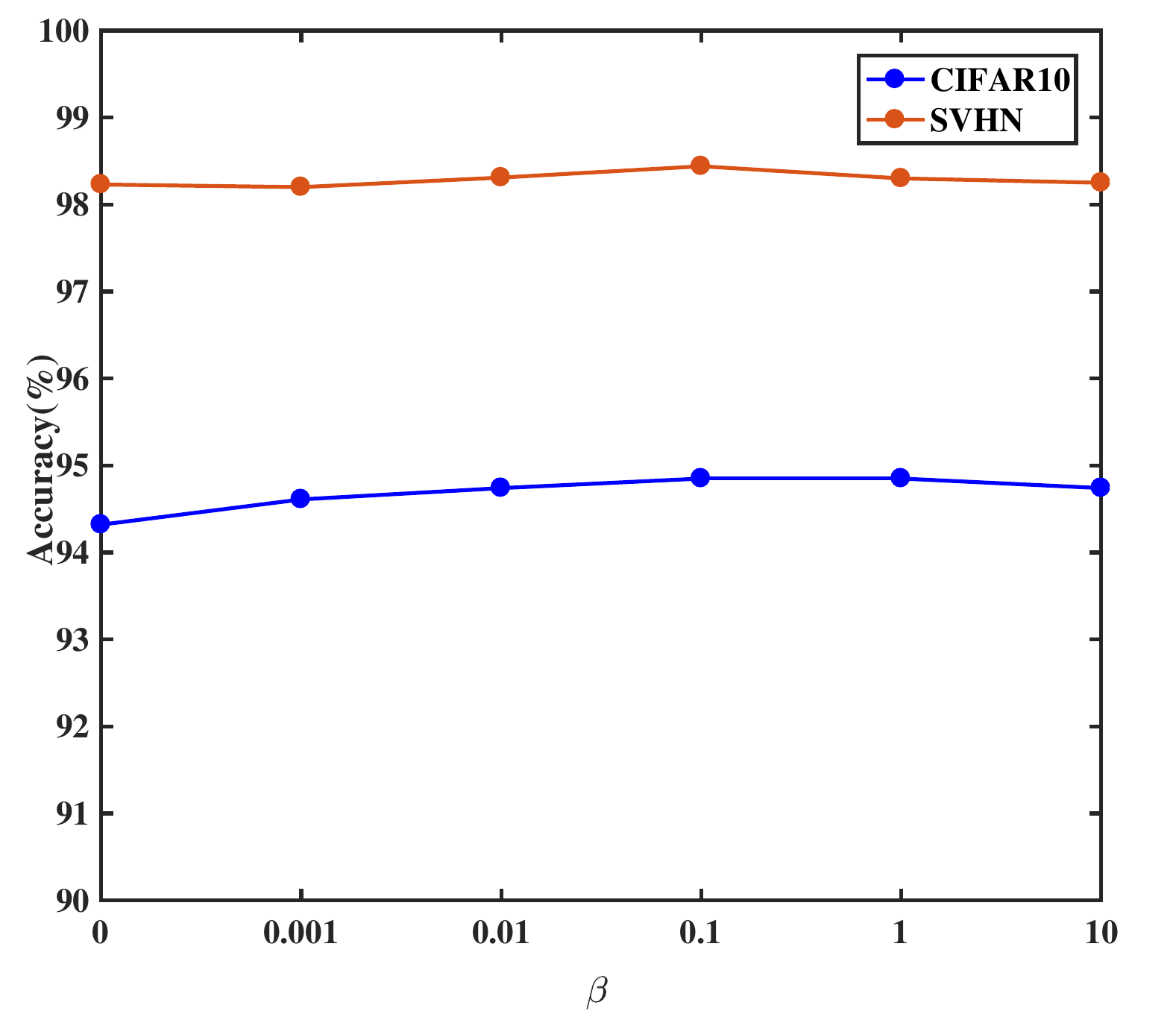}}\caption{Demo of $\beta$ sensitivity in the classification results. Six groups of controlled trials with set $\{0,0.001,0.01,0.1,1,10\}$ are displayed.}\label{fig:beta}} 
\end{figure}

In Equation~\eqref{eq:final-super}, the weight hyper-parameter $\beta$ is introduced to control the weight of the AE regularization term. We further evaluated the susceptibility of $\beta$. Several optional $\beta$ values were pre-specified in the set $\{10,1,0.1,0.01,0.001\}$. Ablation experiments were conducted to understand real impact of the AE regularization term, which corresponds to $\beta=0$. Figure~\ref{fig:beta} shows the classification results on SVHN and CIFAR-10 for demonstration. The results of KLN are insensitive to the value of $\beta$, and the model achieves slightly better performance when $\beta$ is set to 0.1. It is inferred that KLN usually learns the injective mapping with high probability when learning a more discriminative kernel function and parameterizing with deep neural networks.

\begin{table}[h]
\centering
\renewcommand{\tabcolsep}{0.8pc} 
\renewcommand{\arraystretch}{1.2} 
\caption{TRAINING TIME (s) OF CGMMN AND KLN.}\label{tab:runing time}
\begin{tabular}{cccccc}
\hline
				Method		 & MNIST & SVHN &	CIFAR-10	& CIFAR-100\\
\hline	
				CGMMN		 & 26.22  &  453.90   &  37.28	     &43.87\\
				KLN     	 & 62.68 &  907.32   &  75.20	 &81.33 \\
\hline
\end{tabular}
\end{table}

The training time of KLN and CGMMN in one epoch is shown in Table~\ref{tab:runing time}. In principle, the optimization procedure of CMMD contains two separate but related modules, one from the output layer to the latent layer, and the other from the latent layer to the input layer. Both of them can be learned efficiently by the standard Back-Propagation process. Compared with CGMMN, KLN has an additional AE module. Fortunately, the decoder of the AE is mainly composed of de-convolutional operations, which correspond to the convolutional layers used by the encoder. In other words, the AE does not take much time in optimization. It is worth noting that two different subsets were used in each epoch in KLN, while just one subset was used by CGMMN. Thus, KLN costs twice time as much as CGMMN. All the results of the above experiments validate this.

\begin{figure*}[t]
\centering
\subfigure[MNIST with the different labels]{\label{fig:mnist_diff}
\includegraphics[width=2.1in]{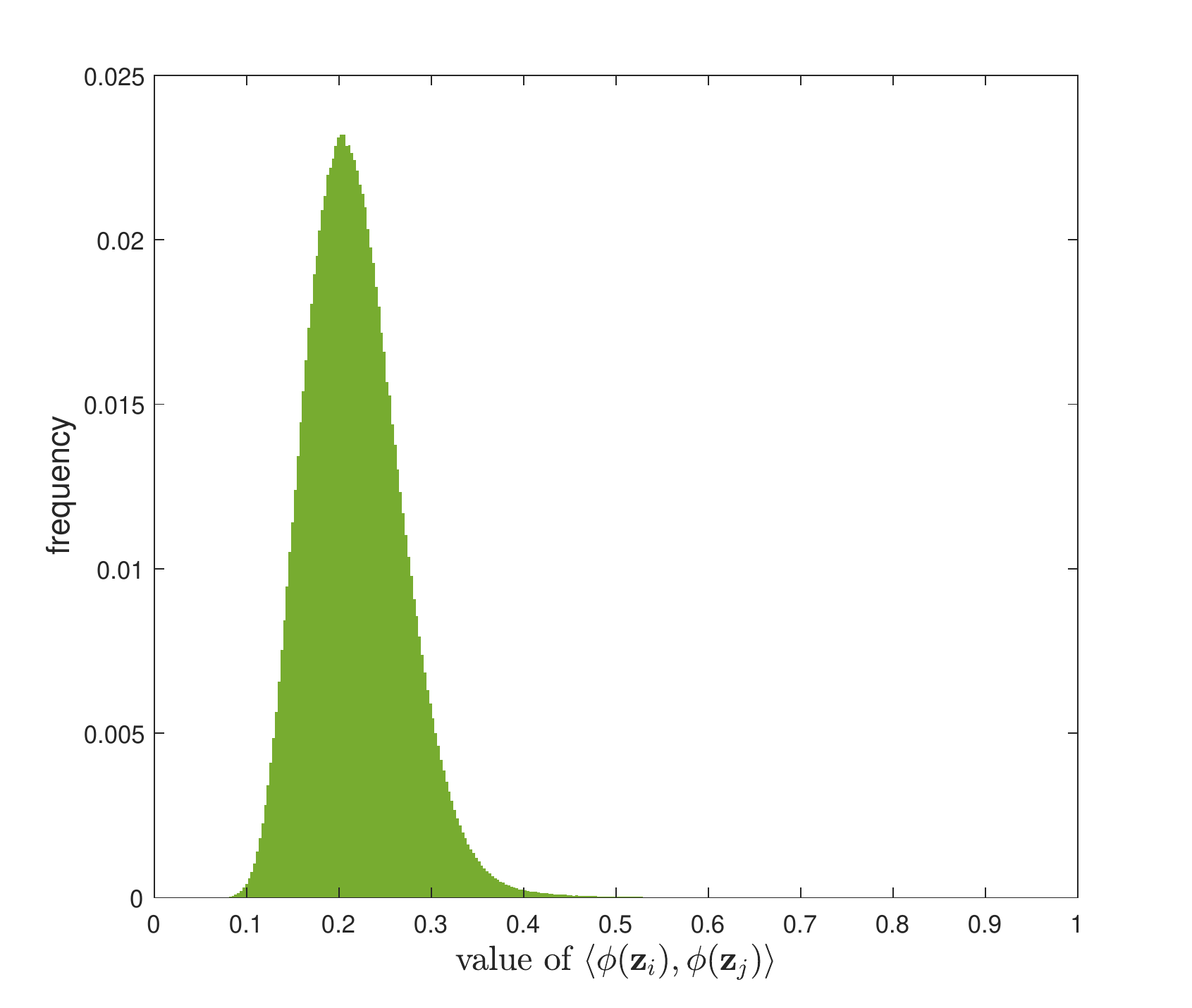}}
\subfigure[SVHN with the different labels]{\label{fig:svhn_diff}
\includegraphics[width=2.1in]{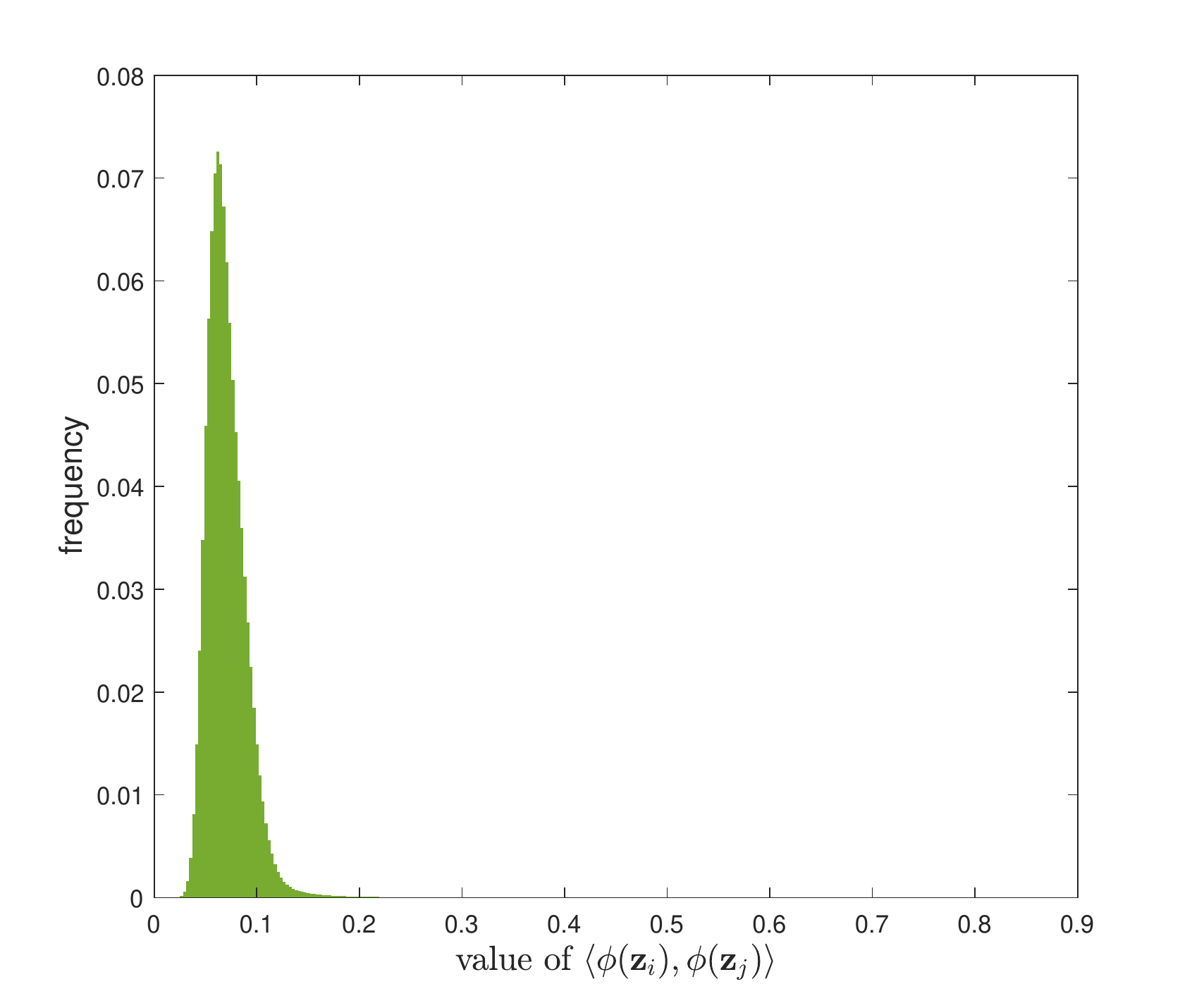}}
\subfigure[CIFAR-10 with the different labels]{\label{fig:cifar_diff}
\includegraphics[width=2.1in]{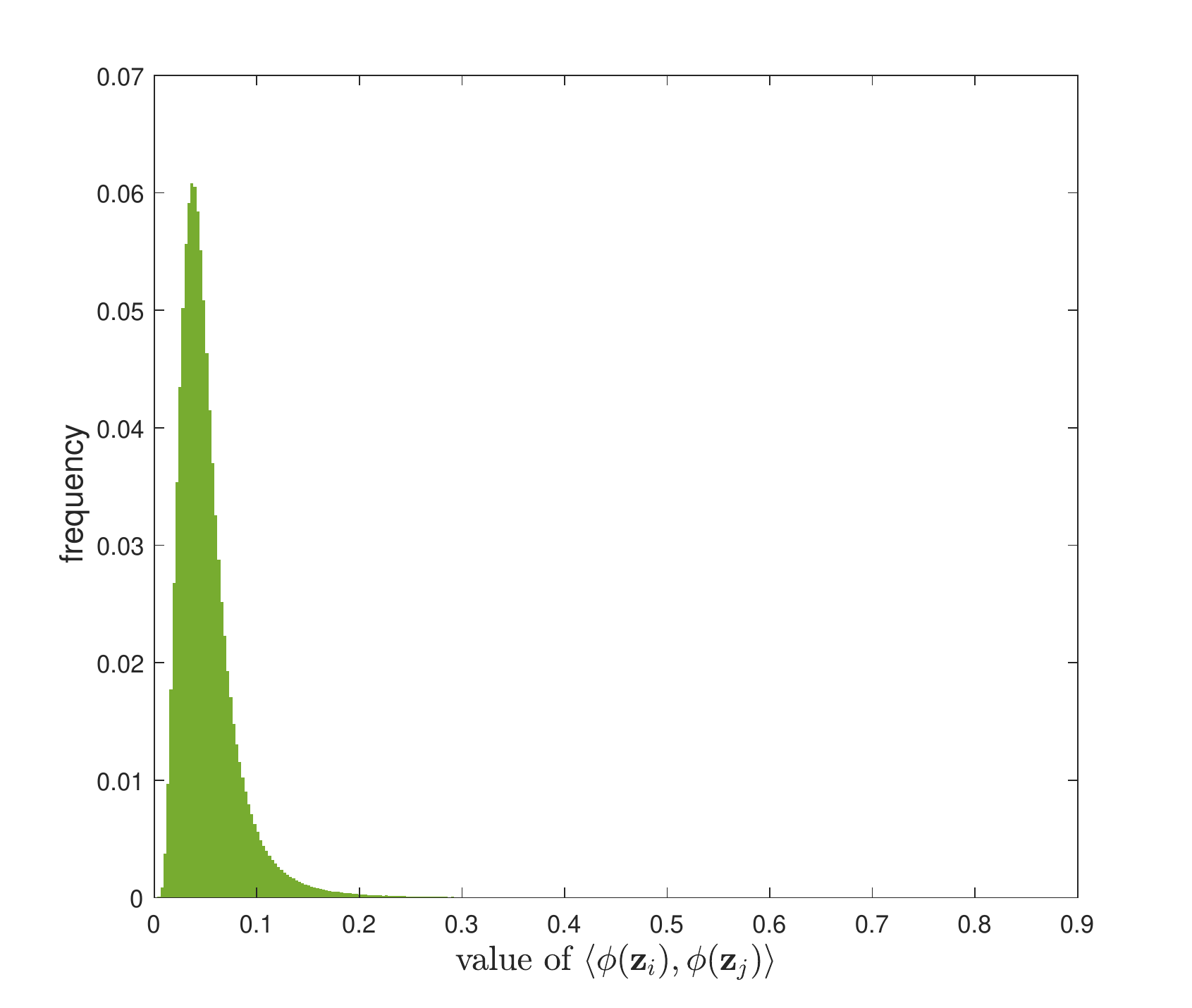}}\\
\subfigure[MNIST with the same label]{\label{fig:mnist_same}
\includegraphics[width=2.1in]{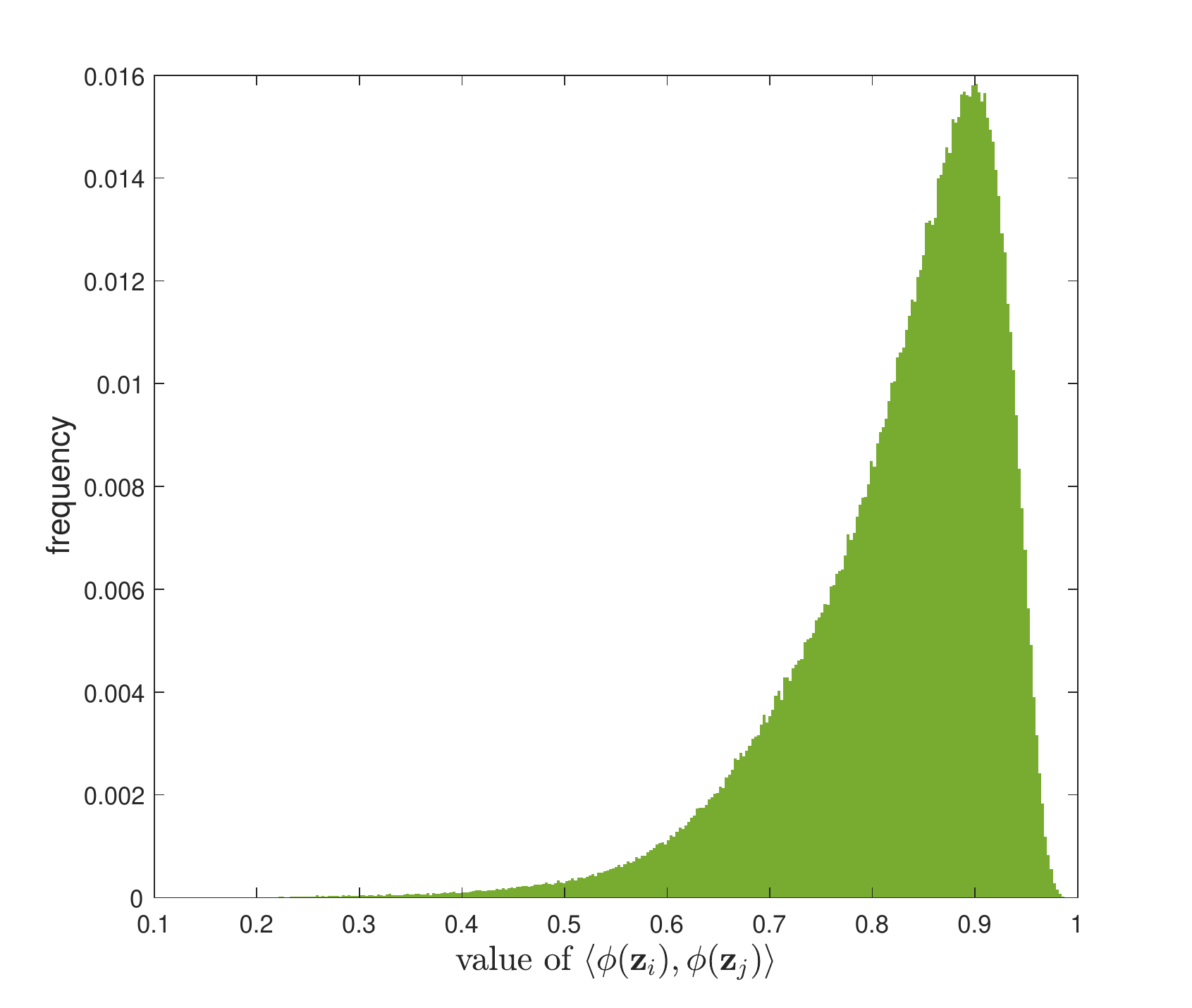}}
\subfigure[SVHN with the same label]{\label{fig:svhn_same}
\includegraphics[width=2.1in]{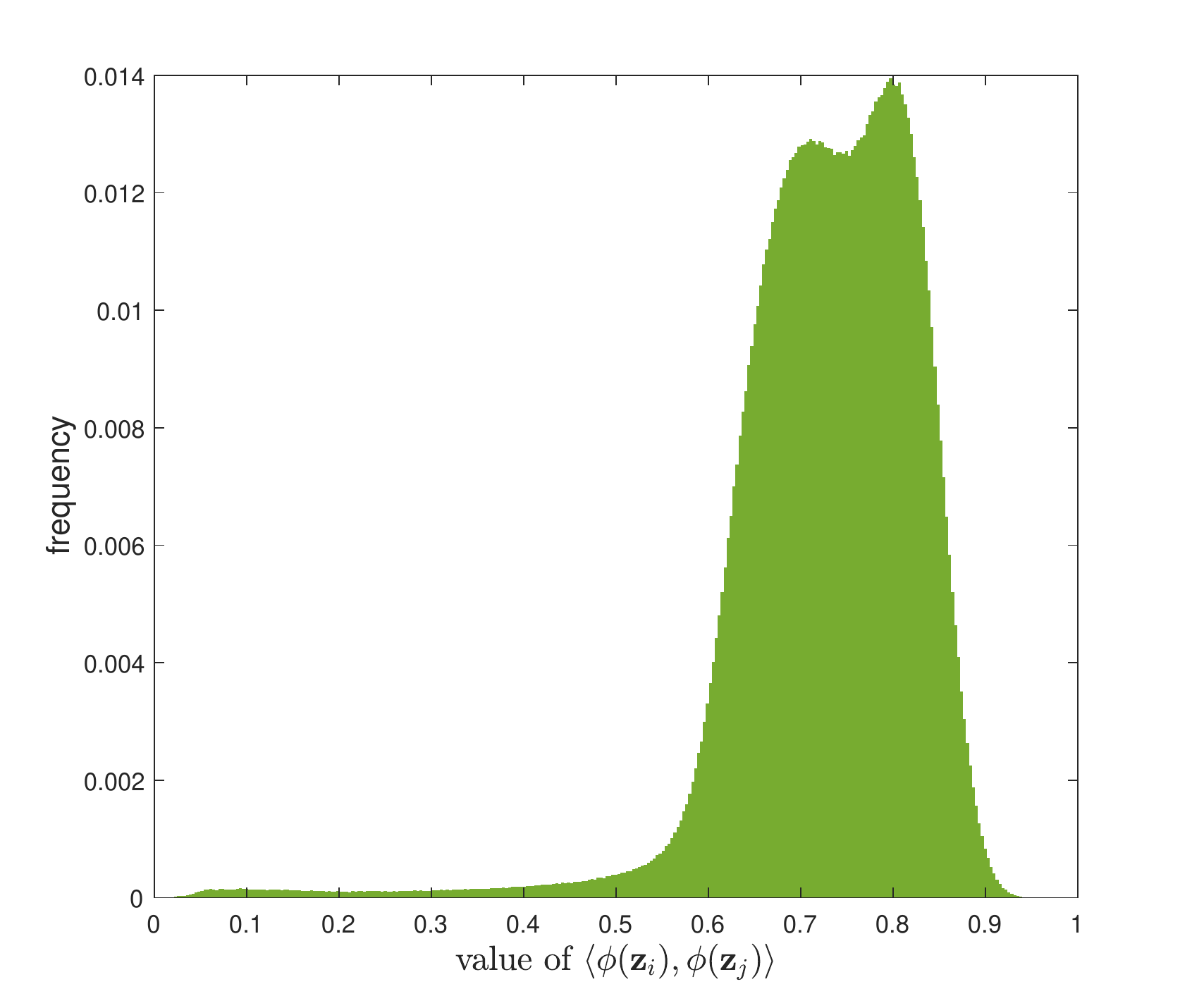}}
\subfigure[CIFAR-10 with the same label]{\label{fig:cifar_same}
\includegraphics[width=2.1in]{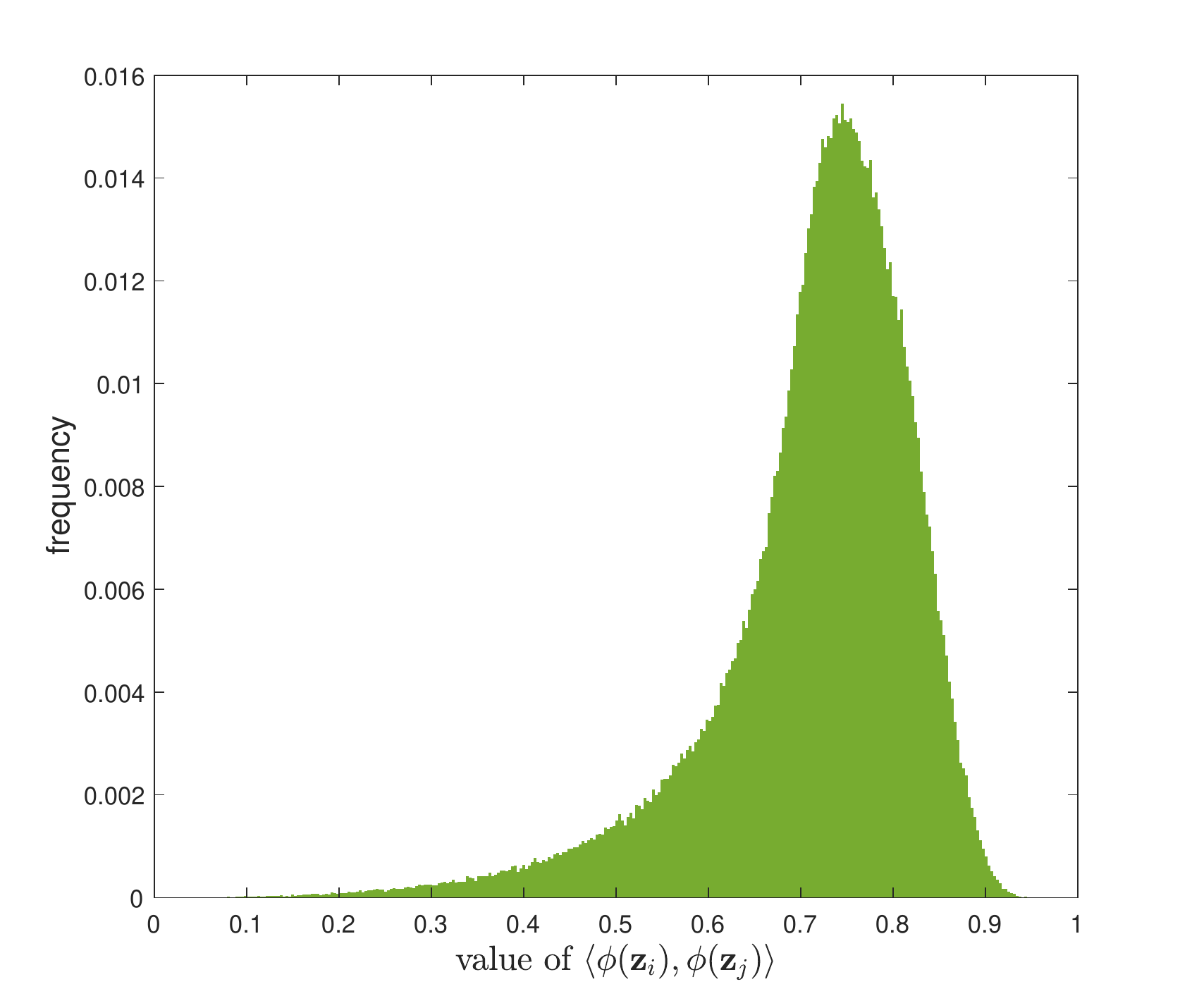}}
\caption{Distribution of $\langle \phi(\textbf{z}_i),\phi(\textbf{z}_j)\rangle$ using samples with same or different labels. It indicates that the kernel values learned by KLN are discriminative and separable.} \label{fig:Zkernel_hist}
\end{figure*}

\subsection{Importance of kernels learning}

In order to evaluate the importance of kernel learning module in KLN, we set up three groups of controlled experiments, which corresponded to three commonly used mapping functions, i.e., identity mapping, PCA and AE respectively, to construct new kernel functions.
\begin{table}[h]
\centering
\renewcommand{\tabcolsep}{0.42pc} 
\renewcommand{\arraystretch}{1.2} 
\caption{Prediction accuracy (\%) of CMMD with different functions $h$ on three benchmark datasets.}\label{tab:results-gcmmn-diff}
\begin{tabular}{cccccc}
\hline
				Dataset		 & Identity Mapping & PCA+CMMD &	AE+CMMD	& KLN\\
\hline
				MNIST		 & 81.62	 &  88.64   & 89.32 	 & \textbf{99.61}\\	
				SVHN		     & 19.59   &  30.23   &  35.23	 & \textbf{98.44}\\
				CIFAR-10	 & 14.25 	 &  27.75   &  33.12	     & \textbf{94.85} \\
\hline
\end{tabular}
\end{table}

A new kernel $\mathcal{K}=k\circ h$, which composes of a mixture of Gaussian kernels and different embedding functions $h$, is considered in this section. When a kernel function is composed with the identity mapping, it means that the kernel matrix is calculated directly from the input data. For the PCA+CMMD and AE+CMMD methods, PCA and AE should be trained additionally, and then the PCA subspace and the encoder of AE are used as the embedding function separately. The critical difference between these methods and KLN is the network learning manner. KLN needs to learn an appropriate kernel function by simultaneously learning a classification model in an end-to-end manner. For a fair comparison, the same network architecture was used to implement these experiments on three datasets. Classification results evaluated on the test sets are shown in Table~\ref{tab:results-gcmmn-diff}.

The classification accuracies of KLN on the three datasets are 99.61\%, 98.44\%, and 94.85\%, respectively. As is shown above, these results are comparable to and even better than those of the state-of-the-art methods. In contrary, corresponding accuracies of using identity mapping are just 81.62\%, 19.59\% and 14.25\%. Obviously, results of KLN are far better than those obtained from identity mapping. It indicates that KLN characterizes the similarity well. The classification accuracies of PCA+CMMD and AE+CMMD on the MNIST are 88.64\% and 89.32\%, respectively, while there are only about 30\% classification accuracies on more complex datasets. It indicates the importance of learning a more effective kernel under complex data distributions.

\subsection{Separating ability of kernels}

In Section~\ref{sect:CGMMN}, we have shown that distributions of $\langle\phi(\textbf{x}_i), \phi(\textbf{x}_j)\rangle$ on samples with the same label or with different labels have no significant difference, thus, the kernel functions used in the CMMD criterion cannot effectively represent similarity. In this paper, a new approach is proposed to extract features via a different kernel utilization manner. Specifically, the new kernel $\mathcal{K}=k\circ h$, in which $k$ is a pre-specified characteristic kernel function and $h$ is the encoder of an auto-encoder network, is exploited in our method. It is expected that the distributions of $\langle\phi(\textbf{z}_i), \phi(\textbf{z}_j)\rangle$ can be well distinguished.

\begin{table*}[htb]
\centering
\renewcommand{\tabcolsep}{1.3pc} 
\renewcommand{\arraystretch}{1.25} 
\caption{Predictive error rates ($\%$) on MNIST, SVHN and CIFAR-10 datasets with $n$ labeled examples, averages of 10 runs.}\label{tab:results-semi}
\begin{tabular}{ccccc}
\hline
				Algorithm		 & MNIST($n=100$) & SVHN($n=1000$) & CIFAR-10($n=4000$) \\
\hline
				DGN~\cite{Kingma2014Semi}		 & 3.33$\pm0.14$	 &  36.02$\pm0.10$   & - \\	
				CatGAN~\cite{Springenberg2016CatGAN}		 & 1.39$\pm0.28$  &   -   &  19.58$\pm0.58$\\
				Ladder network~\cite{Rasmus2015Semi}	 & 1.06$\pm0.37$ &   -   &  20.40$\pm0.47$\\
                ADGM~\cite{Maal2016Auxiliary}		 &  0.96$\pm0.02$ &  22.86    &  -\\	
				SDGM~\cite{Maal2016Auxiliary}		 &  1.32$\pm0.07$	 &  16.61$\pm0.24$    &  -\\
				Improved-GAN~\cite{Salimans2016Improved}	 &  0.93$\pm0.07$	 &  8.11$\pm1.3$    & 18.63$\pm2.32$ \\
                Triple-GAN~\cite{Li2017Triple}	& 0.91$\pm0.58$  & 5.77$\pm0.17$  &
16.99$\pm0.36$     &  \\
                KLN &	 \textbf{0.89$\pm$0.05}  & \textbf{5.51$\pm$0.25}  &
\textbf{16.79$\pm$0.17}     &  \\
\hline
\end{tabular}
\end{table*}

We conducted experiments on three datasets, and then presented histograms of kernel values with the same or different labels in Figure~\ref{fig:Zkernel_hist}. Taking the MNIST dataset for example, Figure~\ref{fig:mnist_diff} shows histogram for the samples with different labels while Figure~\ref{fig:mnist_same} shows histogram for the samples with the same label. It can be seen that the kernel values, i.e., $\langle \phi(\textbf{z}_i), \phi(\textbf{z}_j)\rangle$ of samples with different labels, are mainly concentrated in the range of 0.1-0.3, while those with the same label are concentrated between 0.7 and 1.0. There is a clear distinction between these two
distributions. It indicates that the kernel values learned by KLN have better separating ability. Thus, the effectiveness of KLN in learning more expressive kernel functions is validated.

Results on SVHN and CIFAR-10 present a similar performance. In particular, for SVHN, the kernel values of samples with different labels are mainly concentrated on a sharp neighborhood of 0.0, while those with the same label are concentrated on the range of 0.6-0.9. For CIFAR-10, the kernel values of samples with different labels are mainly concentrated on a sharp neighborhood of 0.1, while those with the same label are concentrated on the range of 0.6-0.9.

We also conducted a group of contrastive experiments to display the histograms of kernel values computed by using the raw input features. The histograms are shown in Appendix~\ref{sect:appendix-histogram}. Specifically, Figures~\ref{fig:kernel_hist_mnist},~\ref{fig:kernel_hist_svhn} and~\ref{fig:kernel_hist_cifar} display the histograms of kernel values on the MNIST, SVHN and CIFAR-10, respectively. All these results indicate that the kernel values, which are directly computed by using the raw features, have little discriminative power in practice.

\begin{figure*}[htb]
\subfigure[Demo of $\beta_1$ sensitivity.]{\label{beta1}
\begin{minipage}[b]{0.5\textwidth} 
\centering \scalebox{0.43}{ 
\includegraphics{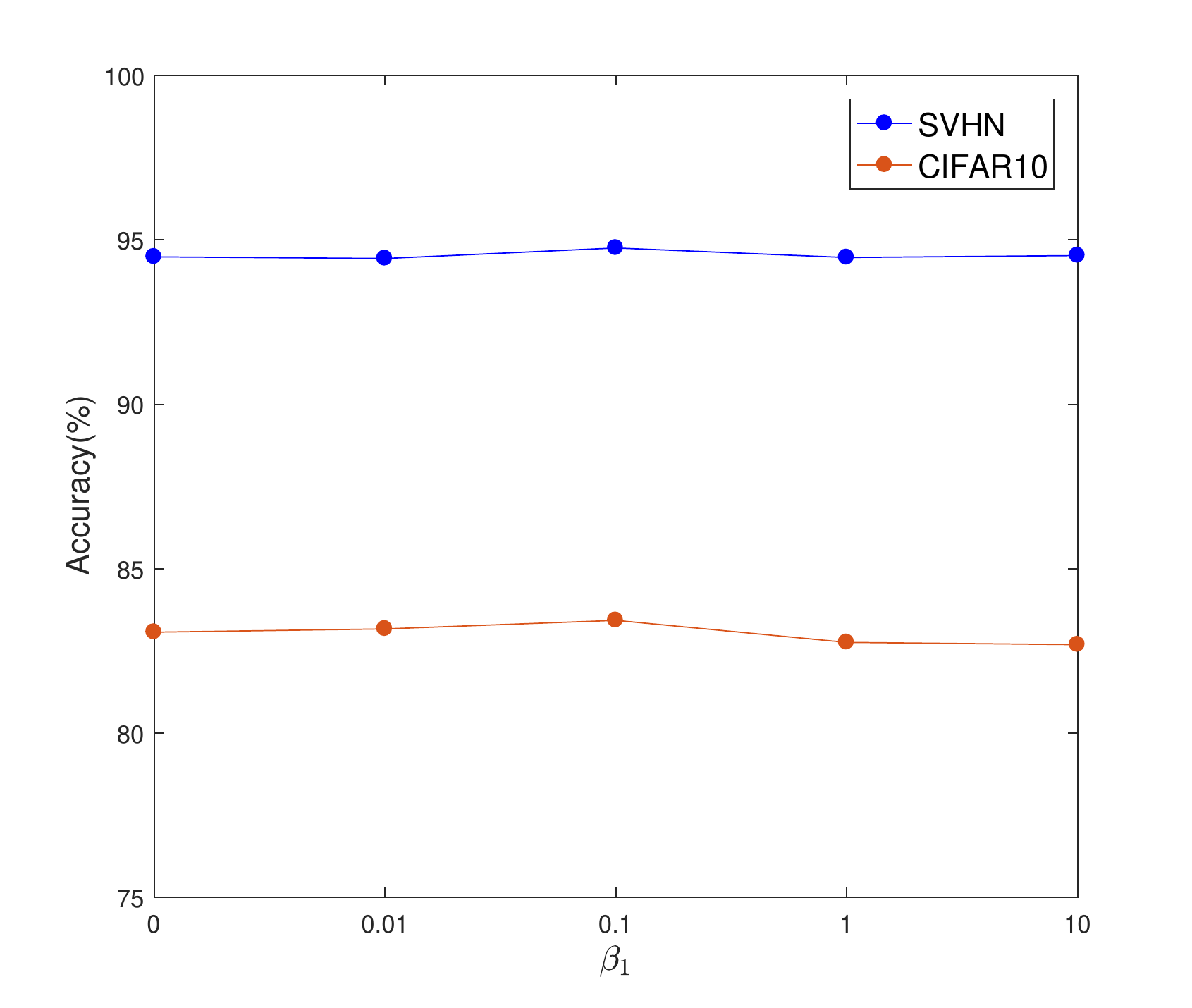}} 
\end{minipage}}
\subfigure[Demo of $\beta_2$ sensitivity.]{\label{beta2}
\begin{minipage}[b]{0.5\textwidth}
\centering \scalebox{0.43}{ 
\includegraphics{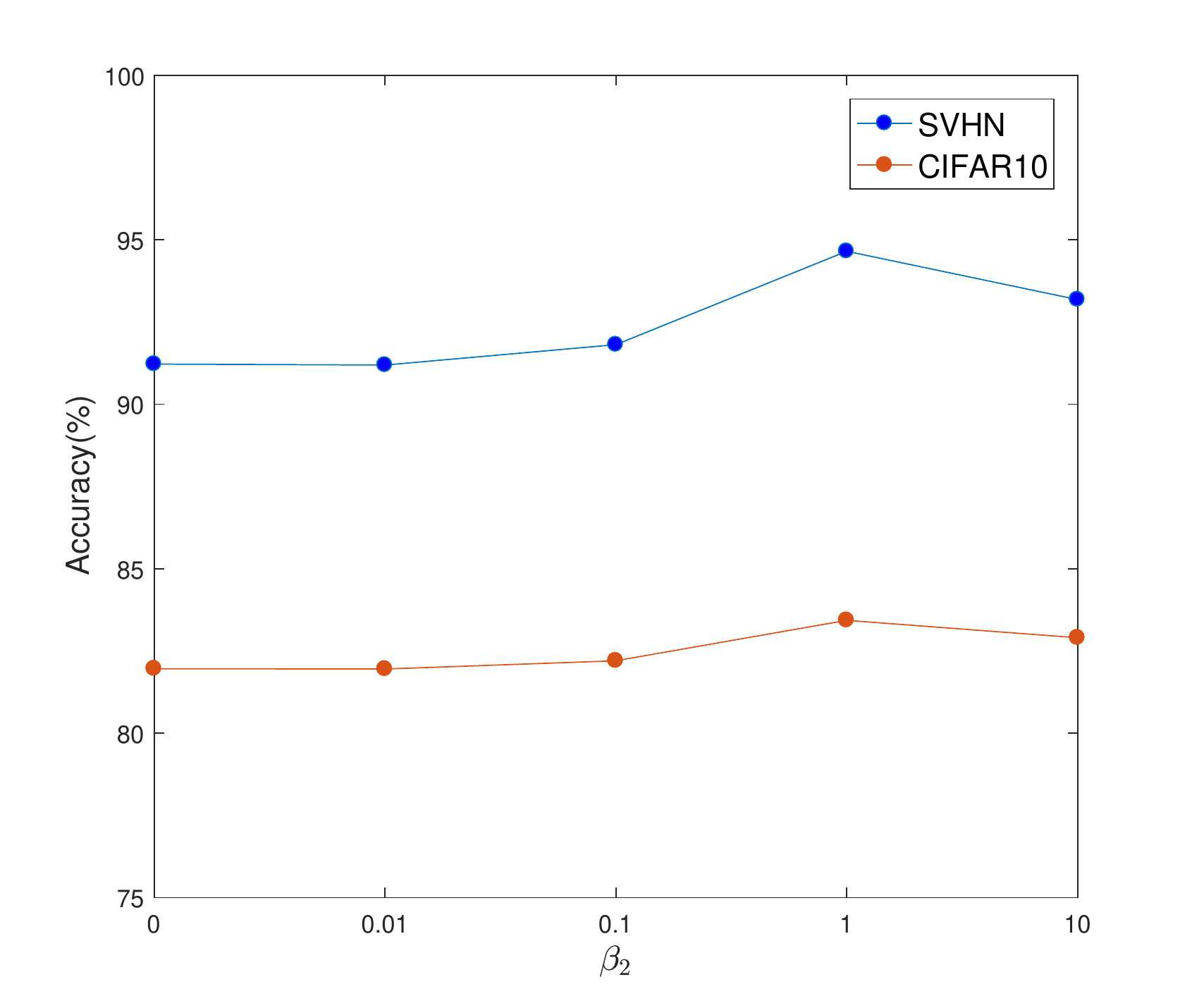}} 
\end{minipage}}
\caption{Demos of $\beta_1$ and $\beta_2$ sensitivity. Five groups of controlled trials with set $\{0,0.01,0.1,1,10\}$ are displayed, respectively. Better viewed in color.}\label{fig:beta12}
\end{figure*}

\subsection{Experiment results of semi-supervised classification}

This section evaluates KLN on semi-supervised classification tasks on MNIST, SVHN and CIFAR-10, and compares the performance with some state-of-the-art semi-supervised methods, including DGN~\cite{Kingma2014Semi}, CatGAN~\cite{Springenberg2016CatGAN}, Ladder network~\cite{Rasmus2015Semi}, ADGM~\cite{Maal2016Auxiliary}, SDGM~\cite{Maal2016Auxiliary}, Improved-GAN~\cite{Salimans2016Improved}, Triple-GAN~\cite{Li2017Triple}.

Following the widely used settings, we used 100, 1000 and 4000 labeled samples on these datasets, respectively. The data preprocessing approaches and network architectures are the same with those used in the supervised learning scenario.

We repeated KLN over 10 runs with different random initializations and random subsets of labeled data, and calculated the mean error rates with the standard deviations following~\cite{Salimans2016Improved}. For a fair comparison, all the results of the baselines are cited from original papers. Table~\ref{tab:results-semi} summarizes the quantitative results. KLN achieves the state-of-the-art predictive accuracies on all these datasets. Note that for a fair comparison with other algorithms, unlike the supervised learning scenario, the extra unlabeled data on SVHN were not used in training.

KLN has two important hyper-parameters, i.e., $\beta_1$ and $\beta_2$, in the semi-supervised learning algorithm. We further evaluated the susceptibility of KLN to these hyper-parameters. These two trade-off parameters were selected from sets $\{10,1,0.1,0.01\}$. Specifically, ablation experiments were conducted to understand the impact of different components in Equation~\eqref{eq:final-semi}, which corresponds to $\beta_1=0$ and $\beta_2=0$, respectively. Figure~\ref{fig:beta12} shows the experiment results on SVHN and CIFAR-10. Parameter $\beta_1$ is the weight of AE regularization term, and the results shown in Figure \ref{beta1} indicate that KLN is insensitive to the value of $\beta_1$. The same phenomenon exists in supervised classification tasks. It can be inferred that KLN usually learns the injective mapping with high probability while learning a more discriminative kernel function and parameterizing with deep neural networks. Parameter $\beta_2$ is the weight of the confidence loss and it could impact the performance of semi-supervised classification significantly. The comparison results are shown in Figure \ref{beta2}. The best choice of $\beta_2$ is 1. When the confidence loss was removed, i.e., $\beta_2=0$, the classification performance dropped but still had competitive results with~\cite{Salimans2016Improved}. It indicates the advantages of KLN and importance of the confidence loss.

\section{Conclusion}\label{sect:conclusion}		

In this paper, image classification methods with the CMMD criterion, which measures difference between two conditional distributions, are studied in depth. It is observed that the classification performance based on CMMD tends to degenerate when the kernel function lacks expressiveness. It means that the kernel functions play an important role in deciding whether classification performance degeneration emerges or not.

The KLN method is proposed to tackle these problems. On one hand, KLN can learn more discriminative kernel functions by minimizing the CMMD criterion and the AE regularization term. On the other hand, KLN can learn kernel functions and label prediction function simultaneously, thus it can be used to deal with classification tasks effectively. KLN is evaluated on four benchmark databases. The results show that it can improve the separating ability of kernel similarities and achieve state-of-the-art performance in both supervised and semi-supervised classification tasks. How to extend KLN to deal with domain adaptation problems is our future work.

\appendices
\section{Heat Maps of Matrix $\textbf{H}$}\label{sect:appendix-heatmap}

\begin{figure}[htb]
\subfigure[]{\label{fig:svhn_C_1}
\begin{minipage}[c]{0.23\textwidth} 
\centering \scalebox{0.25}{ 
\includegraphics{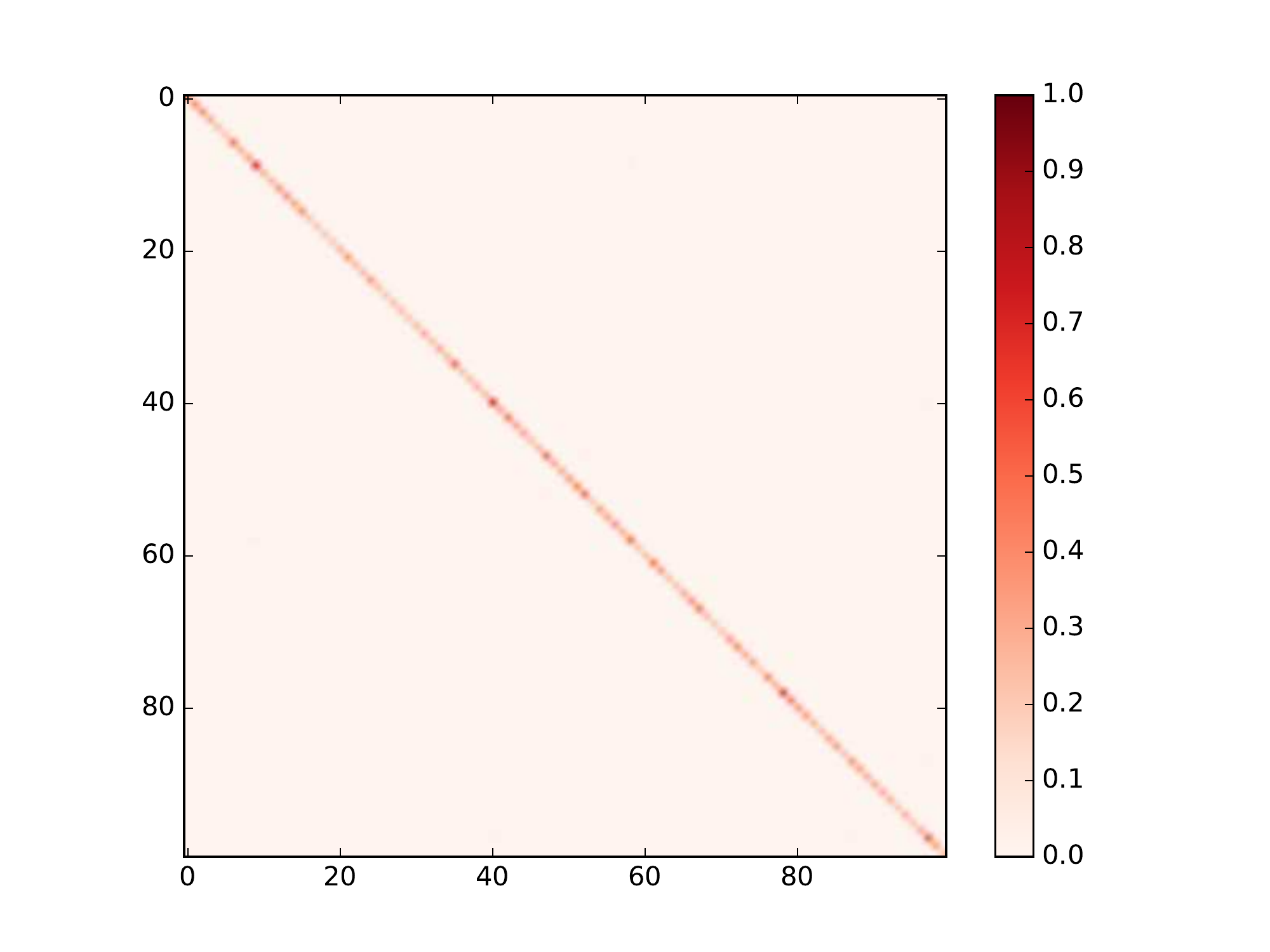}} %
\end{minipage}}
\subfigure[]{\label{fig:ones_svhn_C_1}
\begin{minipage}[c]{0.23\textwidth}
\centering \scalebox{0.25}{ 
\includegraphics{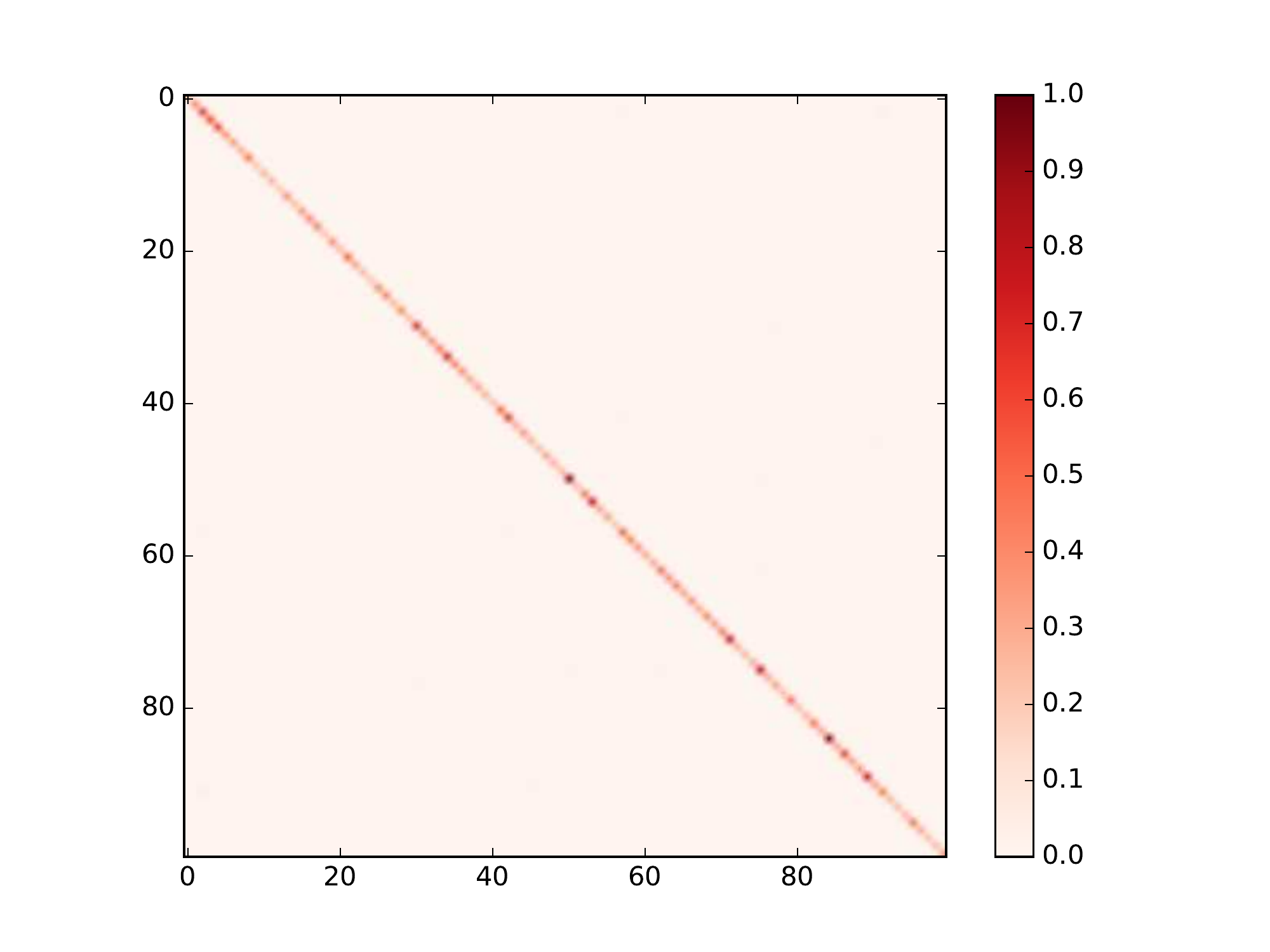}} 
\end{minipage}}
\caption{Heat map of matrix $\textbf{H}$ on SVHN dataset.  (a) The samples were drawn from different classes. (b) The samples were drawn from the same class.}\label{fig:HeatMap-SVHN}
\end{figure}

\begin{figure}[htb]
\subfigure[]{\label{fig:cifar10_C_1}
\begin{minipage}[c]{0.23\textwidth} 
\centering \scalebox{0.25}{ 
\includegraphics{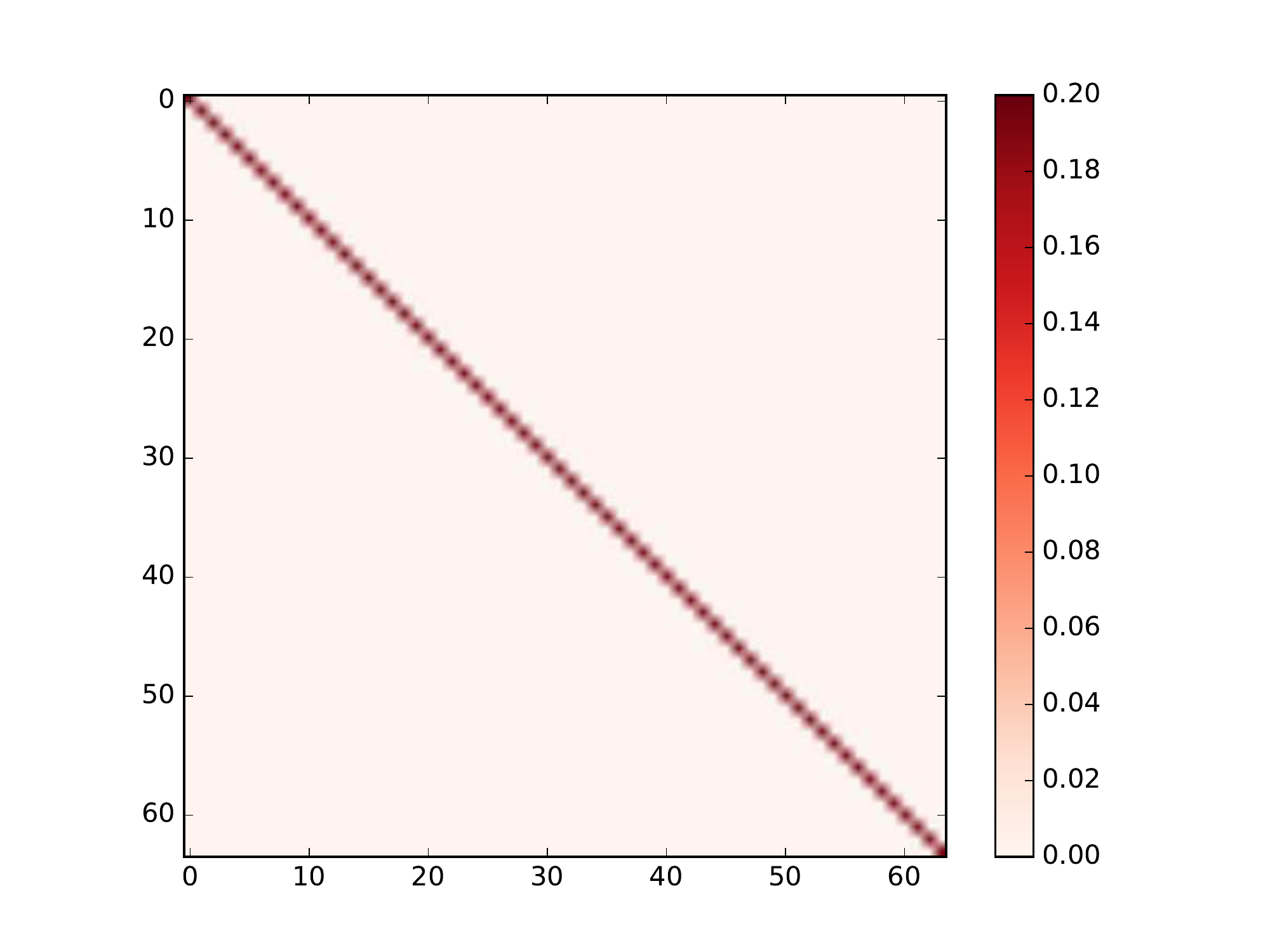}} %
\end{minipage}}
\subfigure[]{\label{fig:ones_cifar10_C_1}
\begin{minipage}[c]{0.23\textwidth}
\centering \scalebox{0.25}{ 
\includegraphics{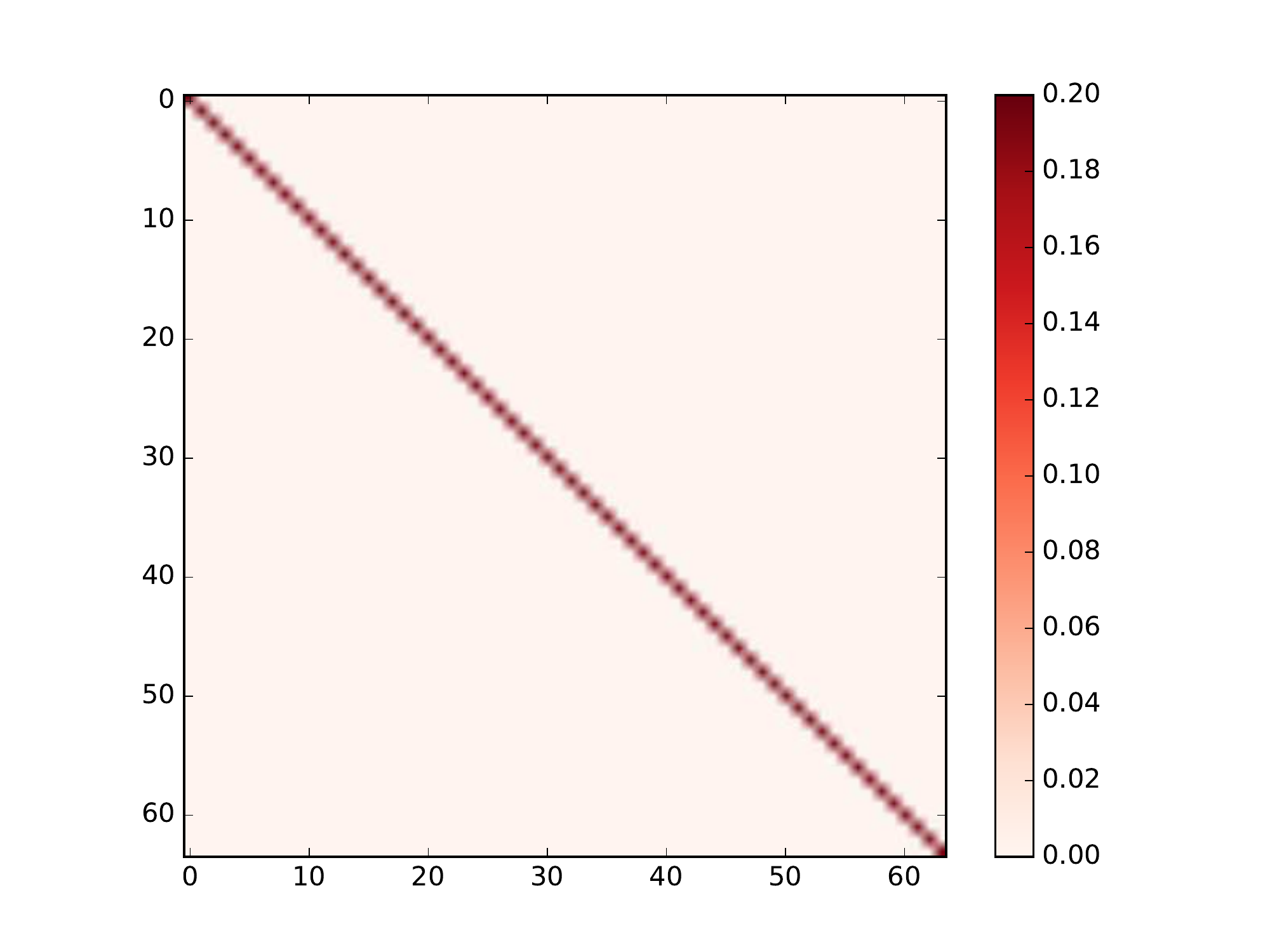}} 
\end{minipage}}
\caption{Heat map of matrix $\textbf{H}$ on CIFAR-10 dataset.  (a) The samples were drawn from different classes. (b) The samples were drawn from the same class.}\label{fig:HeatMap-CIFAR10}
\end{figure}

Figures \ref{fig:HeatMap-SVHN} and \ref{fig:HeatMap-CIFAR10} are the heat maps of matrix $\textbf{H}$ on the SVHN and CIFAR-10 dataset, respectively. The conclusions obtained by these four figures are consistent with the MNIST datasets. Moreover, there is no significant difference between matrices  $\textbf{H}$ for samples of different labels and for samples with the same label.

\vspace{-8pt}

\vspace{-8pt}

\section{Parameter Settings of Models}\label{sect:appendix-network}

The following Table relates to the network structure of KLN in our experiment. The network structure of CGMMN is the same as that of the encoder part of KLN for a fair comparison.
	
The network architecture of encoder was set according to the complexity of the dataset. For the MNIST dataset, the fully convolutional network architecture was used in both encoder and decoder; while for the SVHN and CIFAR-10 datasets, dropout layers and pooling layers were used on the encoder to improve the generalization of the network.

\begin{table}[htb]
		\newcommand{\tabincell}[2]{\begin{tabular}{@{}#1@{}}#2\end{tabular}}
\renewcommand{\tabcolsep}{0.6pc} 
\renewcommand{\arraystretch}{1.1} 
		\begin{center}
			\caption{The network architecture used on MNIST.}
			\label{model_para_mnist}
			\begin{tabular}{|c|c|}
				\hline
				Image size&\tabincell{c}{1$\times$28$\times$28}\\
				\hline			
				Encoder
				&\tabincell{c}
				{$4\times4$ conv, chanel 64, stride 2, padding 1\\
					leakyReLU\\
					$4\times4$ conv, chanel 128, stride 2, padding 1\\
					batch norm, leakyReLU\\
					$7\times7$ conv, chanel 1024, stride 1\\
					batch norm, leakyReLU\\
					$1\times1$ conv, chanel 128, stride 1\\
					batch norm, leakyReLU				
				}\\
				\hline
				Dimension of $\textbf{Z}$	&128\\
				\hline	
				Decoder
				&\tabincell{c}{
					$1\times1$ deconv, chanel 1024, stride 1\\
					batch norm, ReLU\\
					$7\times7$ deconv, chanel 128, stride 1\\
					batch norm, ReLU\\
					$4\times4$ deconv, chanel 64, stride 2, padding 1\\
					batch norm, ReLU\\
					$4\times4$ deconv, chanel 1, stride 2, padding 1\\
					sigmoid
				}\\	
				\hline
				$\sigma^2$	&\{1,3,5,7,9\}\\
				\hline
			\end{tabular}
		\end{center}
\end{table}

\section{Histogram of the kernel similarities}\label{sect:appendix-histogram}

\begin{figure}[htb]
\subfigure[]{\label{fig:mnist_dist}
\begin{minipage}[c]{0.23\textwidth}
\centering \scalebox{0.25}{
\includegraphics{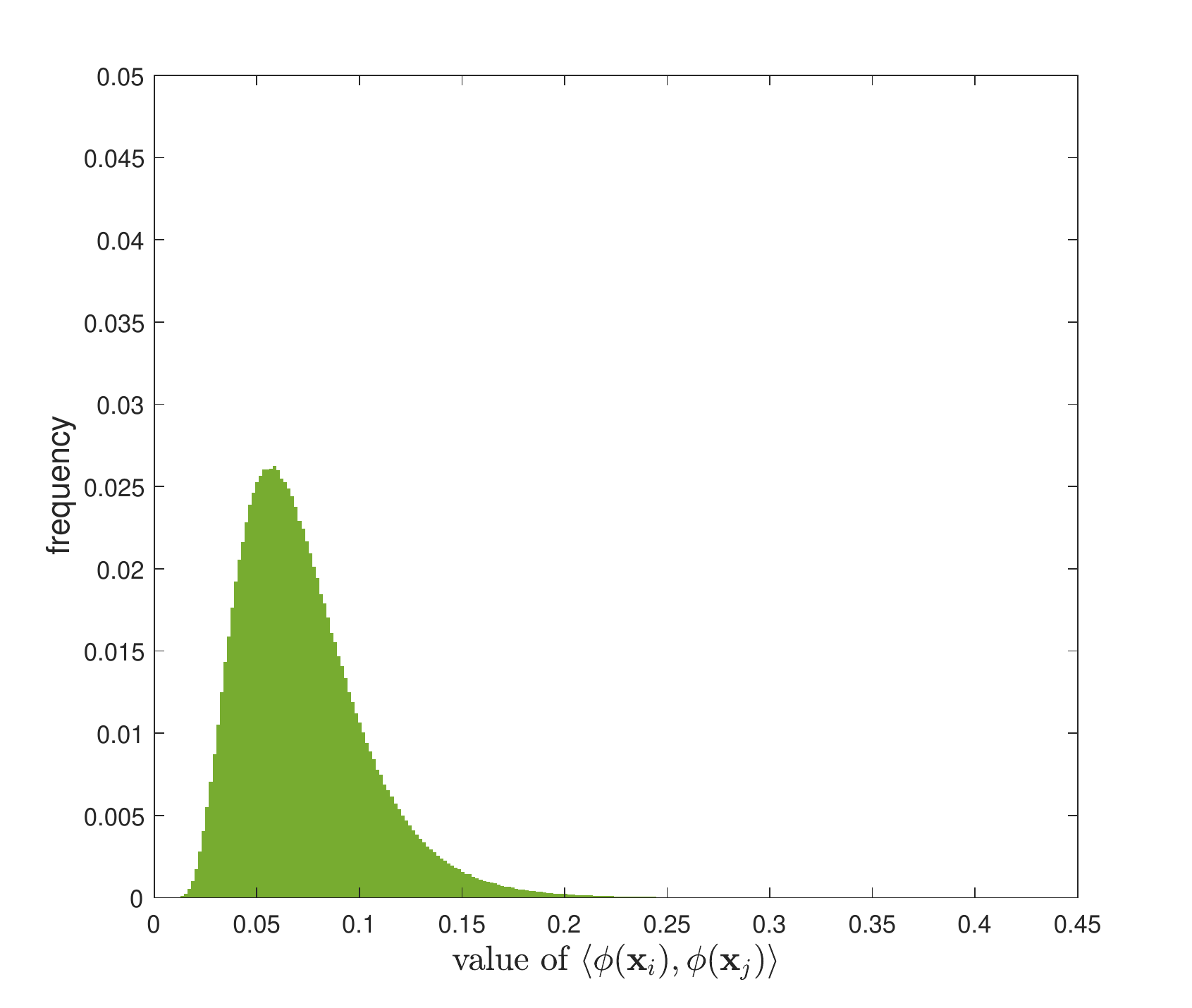}}
\end{minipage}}
\subfigure[]{\label{fig:ones_mnist_dist}
\begin{minipage}[c]{0.23\textwidth}
\centering \scalebox{0.25}{
\includegraphics{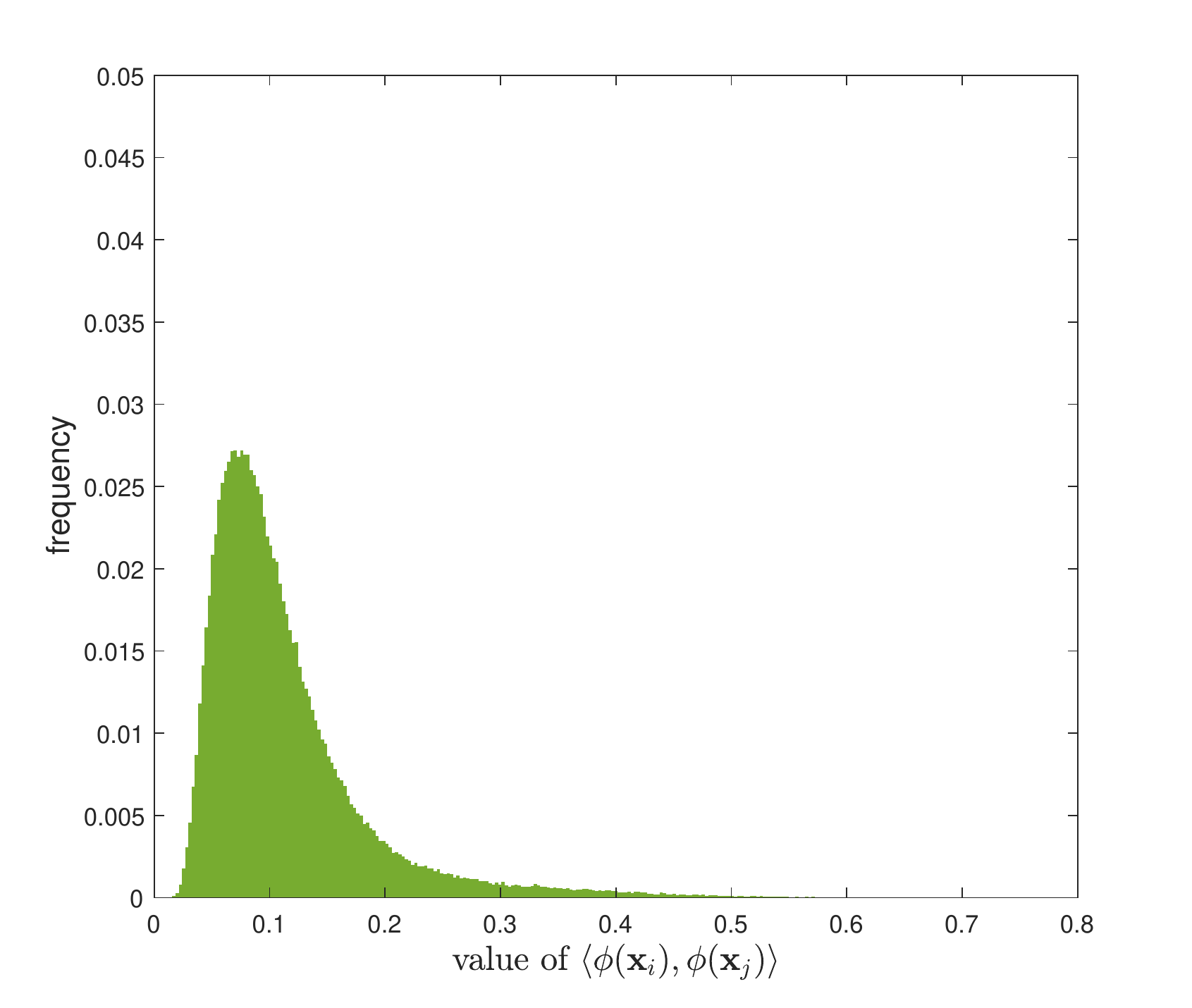}}
\end{minipage}}
\caption{Distribution of $\langle \phi(\textbf{x}_i),\phi(\textbf{x}_j)\rangle$ on MNIST. (a) The samples were drawn from different classes. (b) The samples were drawn from the same class.}\label{fig:kernel_hist_mnist}
\end{figure}

\begin{figure}[htb]
\subfigure[]{\label{fig:svhn_dist}
\begin{minipage}[c]{0.23\textwidth}
\centering \scalebox{0.25}{
\includegraphics{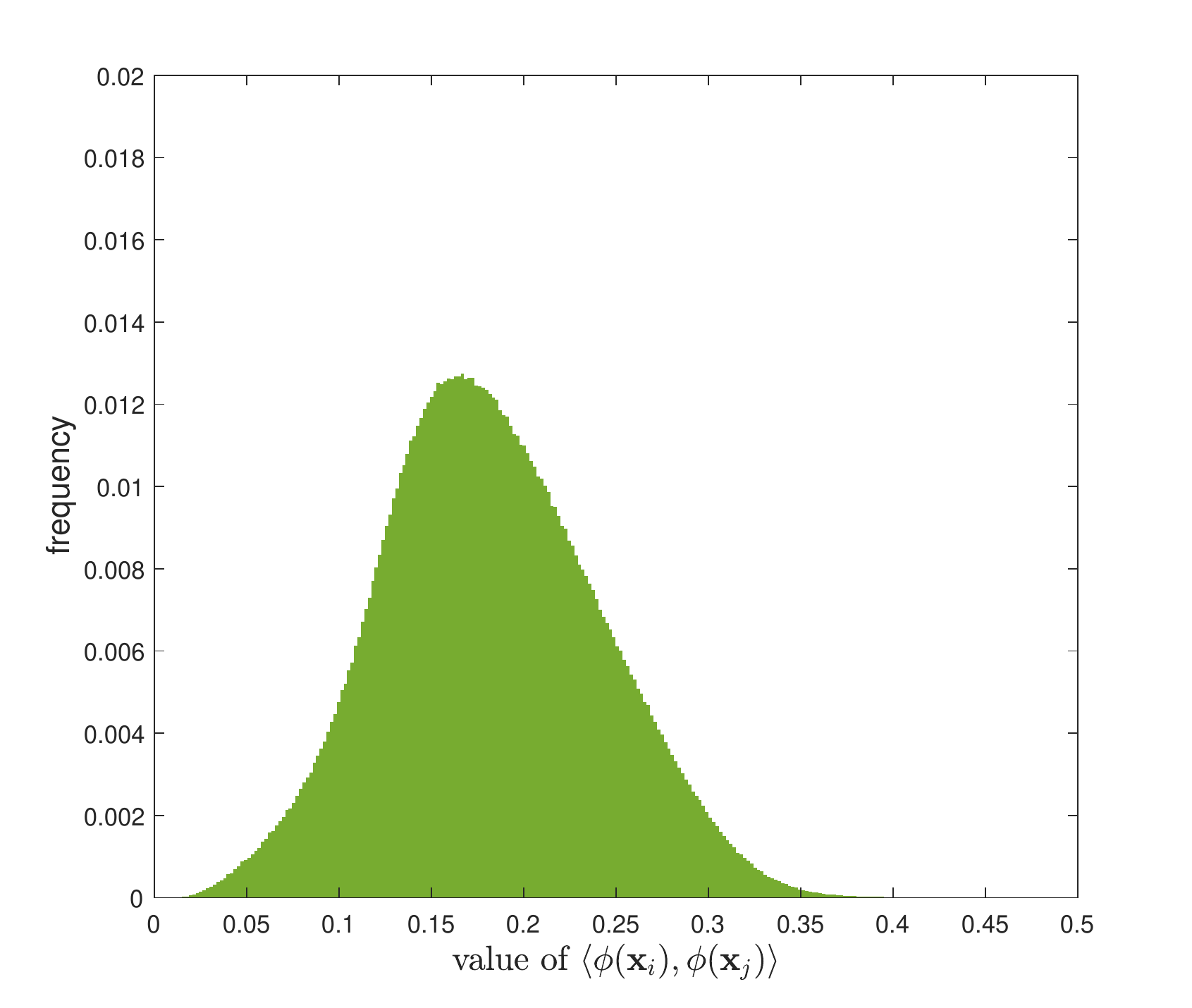}}
\end{minipage}}
\subfigure[]{\label{fig:ones_svhn_dist}
\begin{minipage}[c]{0.23\textwidth}
\centering \scalebox{0.25}{
\includegraphics{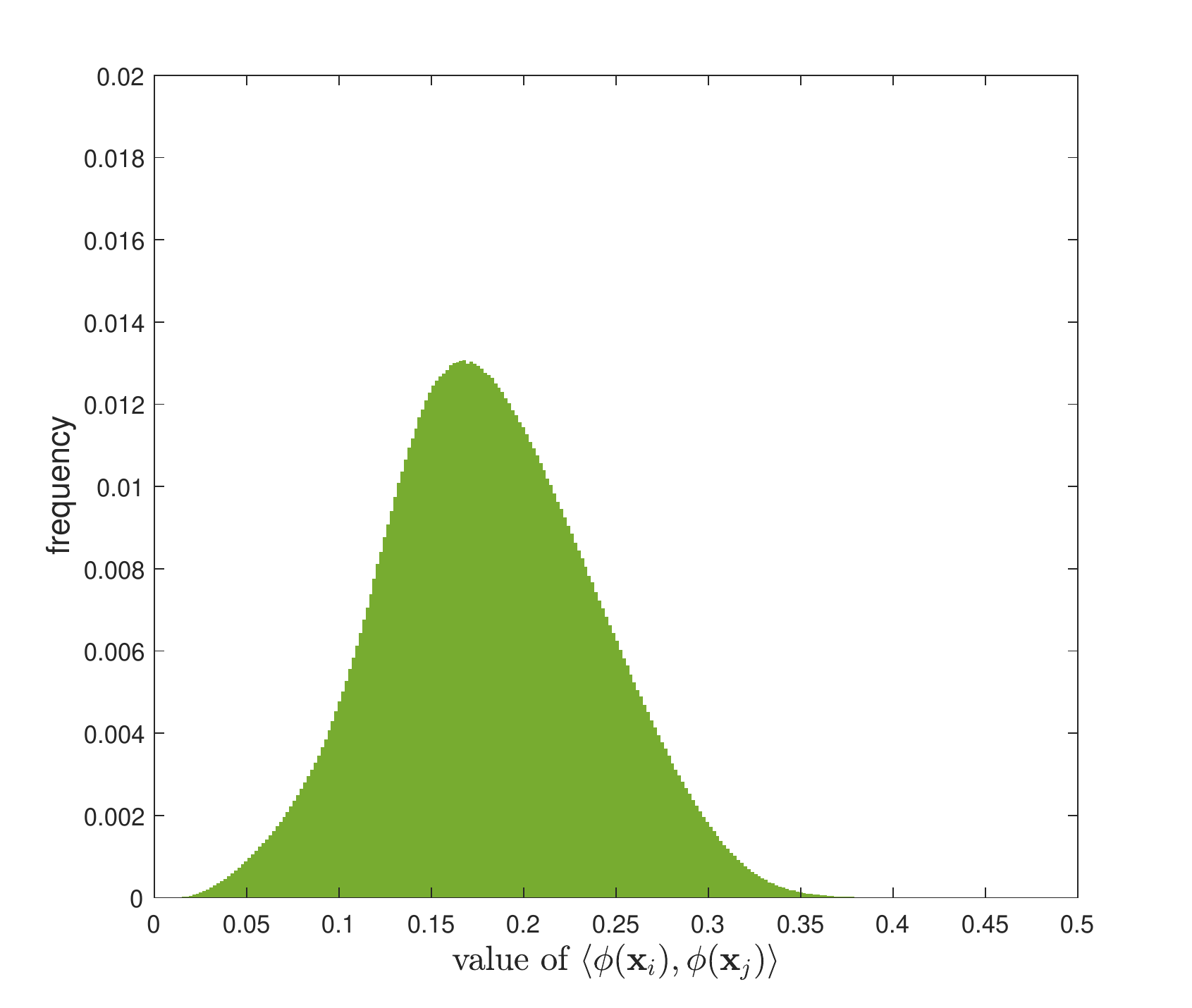}}
\end{minipage}}
\caption{Distribution of $\langle \phi(\textbf{x}_i),\phi(\textbf{x}_j)\rangle$ on SVHN.  (a) The samples were drawn from different classes. (b) The samples were drawn from the same class.}\label{fig:kernel_hist_svhn}
\end{figure}

\begin{figure}[!h]
\subfigure[]{\label{fig:cifar_dist}
\begin{minipage}[c]{0.23\textwidth}
\centering \scalebox{0.25}{
\includegraphics{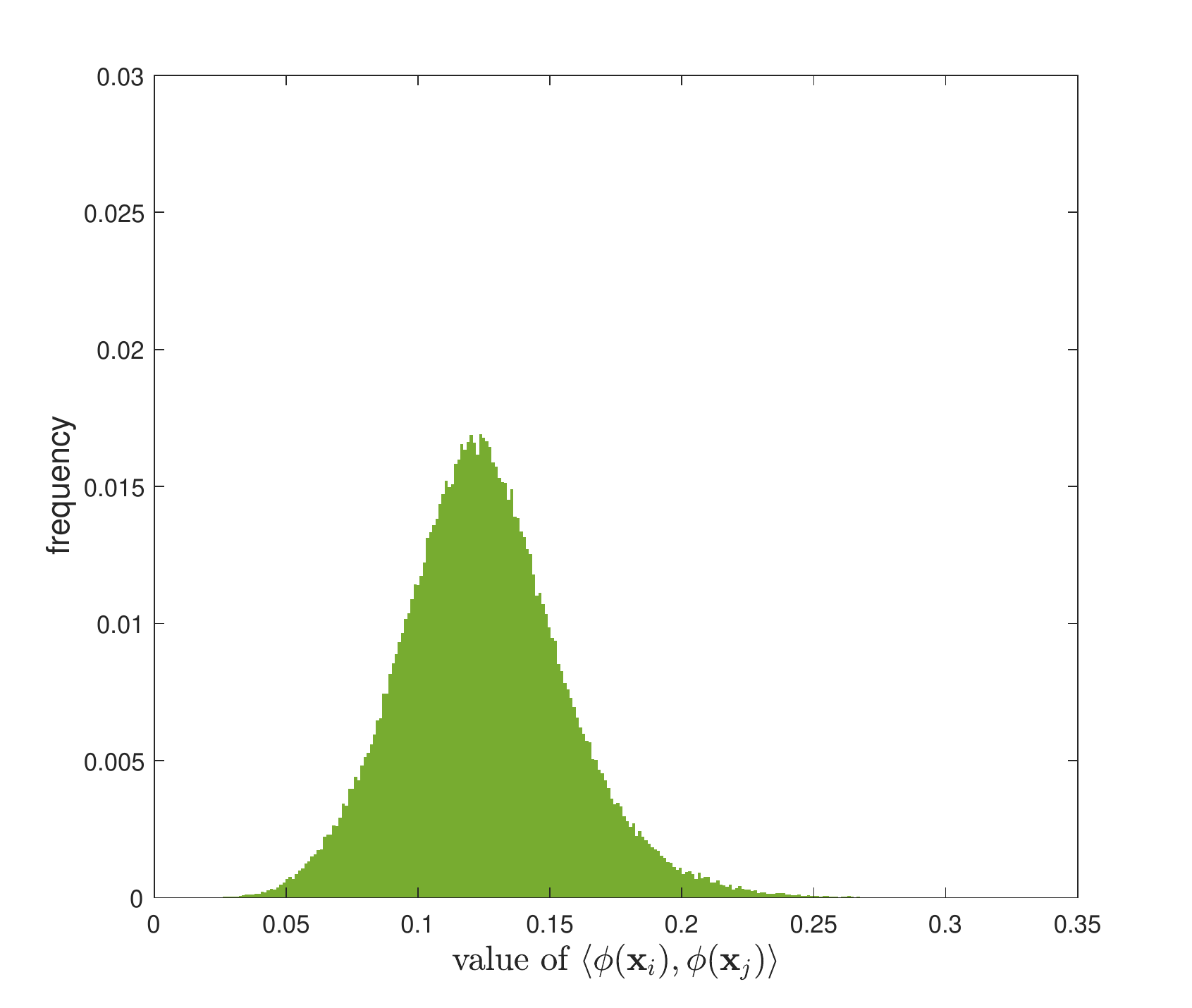}}
\end{minipage}}
\subfigure[]{\label{fig:ones_cifar_dist}
\begin{minipage}[c]{0.23\textwidth}
\centering \scalebox{0.25}{
\includegraphics{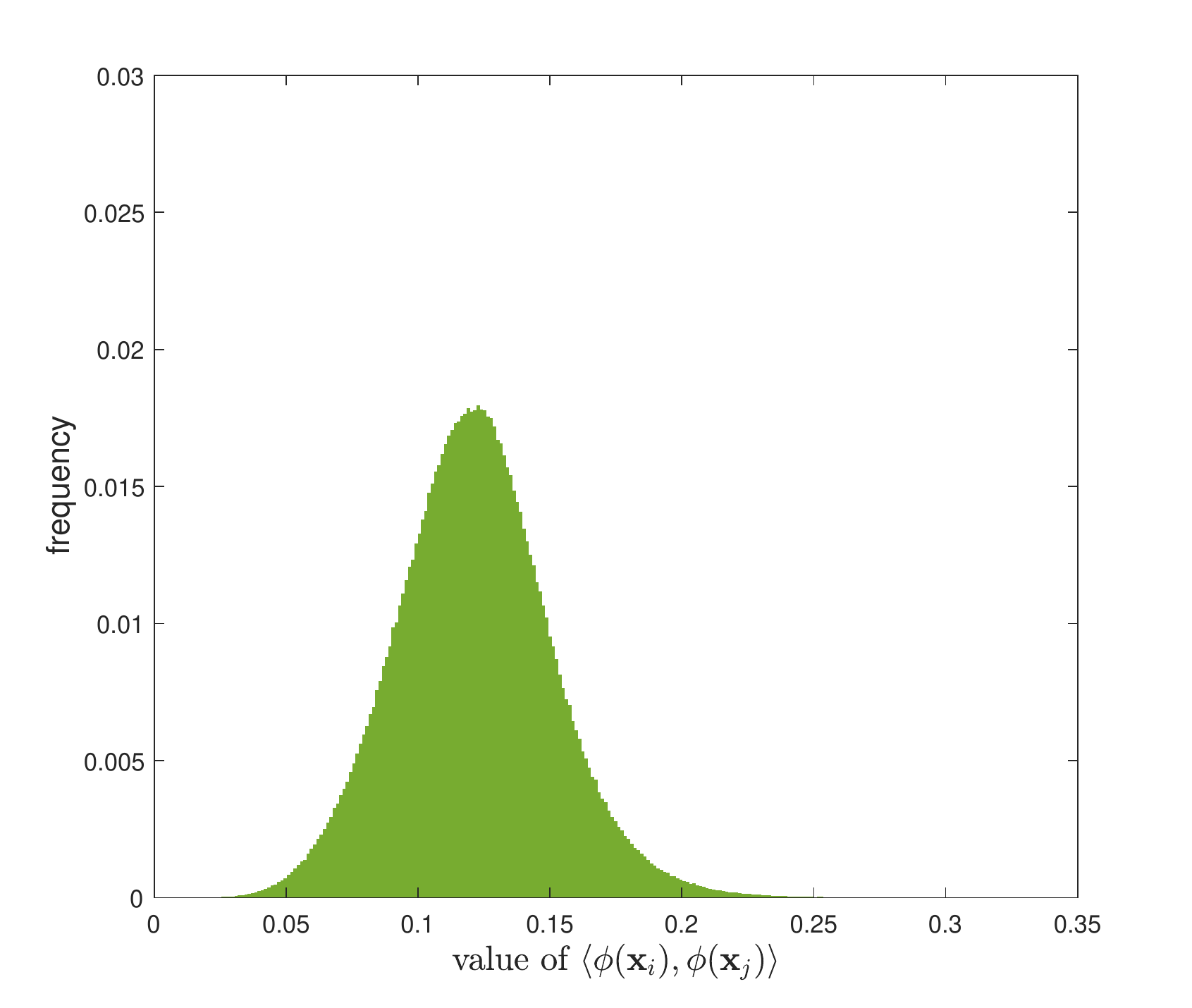}}
\end{minipage}}
\caption{Distribution of $\langle \phi(\textbf{x}_i),\phi(\textbf{x}_j)\rangle$ on CIFAR-10.  (a) The samples were drawn from different classes. (b) The samples were drawn from the same class.}\label{fig:kernel_hist_cifar}
\end{figure}

Figures~\ref{fig:kernel_hist_mnist},~\ref{fig:kernel_hist_svhn} and~\ref{fig:kernel_hist_cifar} display the histograms of kernel values computed by using the raw input features, on the MNIST, SVHN and CIFAR-10, respectively. We can see that the histogram distributions of between-classes (the left diagram) and within-classes (the right diagram) have no significant difference.

\bibliographystyle{IEEEtran}
\bibliography{KLN}

\end{document}